# Microelectronic Morphogenesis: Progress towards Artificial Organisms

*John S. McCaskill[1,2,3*], Daniil Karnaushenko[1,2*], Minshen Zhu[1,2] and Oliver G. Schmidt[1,2,3*]*

[1] Research Center for Materials, Architectures and Integration of Nanomembranes (MAIN), Chemnitz University of Technology, 09126 Chemnitz, Germany
[2] Material Systems for Nanoelectronics, Chemnitz University of Technology, 09126 Chemnitz, Germany
[3] European Centre for Living Technology (ECLT), Ca' Bottacin, Dorsoduro 3911, Venice, 30123, Italy

*Corresponding authors e-mail: john.mccaskill@main.tu-chemnitz.de ; daniil.karnaushenko@main.tu-chemnitz.de ; oliver.schmidt@main.tu-chemnitz.de

## Abstract

Microelectronic morphogenesis is the creation and maintenance of complex functional structures by microelectronic information within shape-changing materials. Only recently has in-built information technology begun to be used to reshape materials and their functions in three dimensions to form smart microdevices and microrobots. Electronic information that controls morphology is inheritable like its biological counterpart, genetic information, and is set to open new vistas of technology leading to artificial organisms when coupled with modular design and self-assembly that can make reversible microscopic electrical connections. Three core capabilities of cells in organisms, *self-maintenance* (homeostatic metabolism utilizing free energy), *self-containment* (distinguishing self from non-self), and *self-reproduction* (cell division with inherited properties), once well out of reach for technology, are now within the grasp of information-directed materials. Construction-aware electronics can be used to proof-read and initiate game-changing error correction in microelectronic self-assembly. Furthermore, non-contact communication and electronically supported learning enable one to implement guided self-assembly and enhance functionality. This article reviews the fundamental breakthroughs that have opened the pathway to this prospective path, analyzes the extent and way in which the core properties of life can be addressed and discusses the potential and indeed necessity of such technology for sustainable high technology in society.

## 1. Introduction

*Living technology* – technology invested with the information-controlled sustainable self-x properties of living systems, like self-organization, self-maintenance, self-repair, and self-awareness – proposed two decades ago[1], has in the past few years leapt towards realization *via* a new generation of microrobotic flexible electronics. This review covers the recent developments that now clearly project towards this goal, one that is becoming more urgent for sustainability as the tasks, density, and impact of technology in our world dramatically increase. While macroscopic humanoid programmable robots like Ameca are also becoming steadily more life-like[2], they lack the microscopically modular information control of shape-changing structures that cells provide to organisms, allowing the latter to grow and self-repair using a flux of resources. At molecular nanoscales, although progress in chemistry, functional materials and synthetic biology fills many journals, the development of nanorobotics is either with a low level of information control (as in smart nanoparticles for medicine[3]) or bound closely to the molecular constraints of biological organisms as in artificial DNA nanomachines[4]. Where we have independently achieved dense information processing, based on 2D patterning as in silicon chip CMOS technology, until recently it has been only with rigid fixed structures with little or no inbuilt information control over their shapes or three-dimensional structures let alone over shape dynamics. Macroscopic modular robotics[5] makes use of silicon chips to control complex machine modules able to self-assemble and change their shapes, but the large size (and cost) of modules limits their ability to exploit self-x properties. Soft modular robotics has been a major focus of recent developments[6–9] but integration with electronic control has lagged. This review addresses the advances in technology that are enabling a radical downscaling of modular robotics to use smart microscopic modules with on board microelectronics, and in particular advances that are projecting such systems towards a low module size, high flexibility and high information-control sub-domain in which *microelectronic morphogenesis* becomes possible: the generation of organism-like structural complexity through the self-folding and self-assembly of microelectronically active micromodules.

The fascination of self-organization of living organisms and inanimate processes and machines that capture even a small fraction of life's potential is apparent in human history and recent developments: from the mastery of fire to steam engine thermodynamics[10] and life as a non-equilibrium process[11]; from chemical morphogenesis[12],[13] to mechanochemical pattern formation[14] and from von Neumann's self-reproducing automata to mechanical self-reproduction[15][16]; from Leonardo da Vinci's flying machines[17] and Rechenberg's evolving wings[18] to biomimetics[19,20], living technology[21], artificial cells[22,23], soft robotics[24], evolution of new materials[25–27], evolvable hardware[28,29] and evolutionary robotics[30]. Inside cells, complex biological machines assemble, maintain, and copy themselves, performing multi-scale tasks with precision and reliability, and evolving sustainably into an almost incomprehensible variety of structure and function over millions of years. Organisms consist of cells with three core capabilities[31], which are key to their persistence in the world: *self-maintenance* (homeostatic metabolism utilizing free energy), *self-containment* (formation of a boundary, distinguishing self from non-self), and *self-reproduction* (cell division with inherited properties). Controlled locomotion is sometimes seen as fundamental also to achieving self-reproduction and hence a core capability of cells[32]. In addition, many cells can change shape and move, sense their environment, communicate with one another, self-assemble into larger systems and act collectively. To do all this, cells deploy a common architectural principle of self-assembly of the folded flexible biopolymer products of genetically encoded processes.

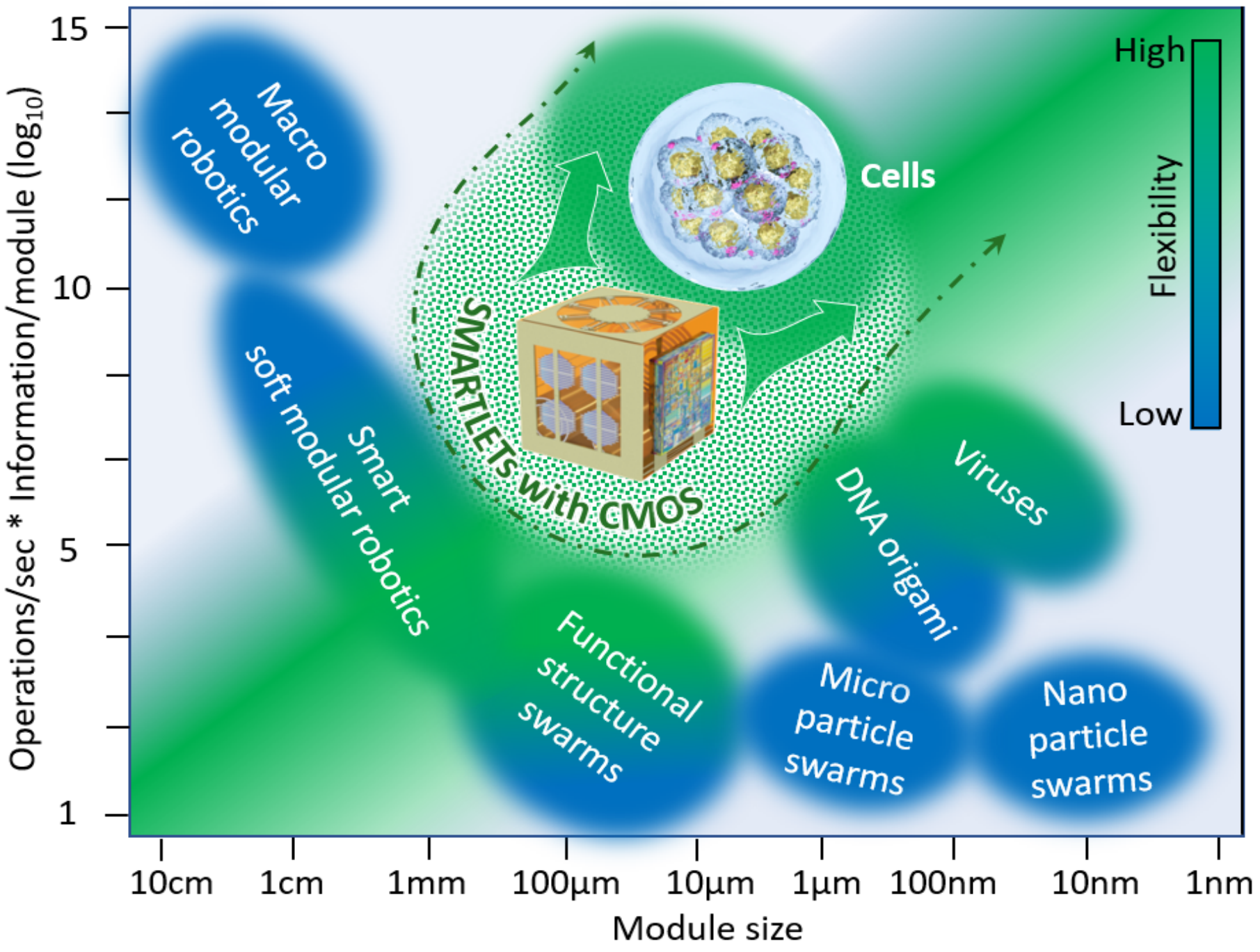

**Figure 1 Characterization of various modular functional technologies with respect to living systems on three axes:** size, amount of information processing, and shape flexibility. For size we employ the largest dimension in the completely folded state. For information processing we take into account both the amount of information per module and the number of specific operations performed per second. For functional flexibility in shape-changing we use the % of size that the resting shape can be strained to implement a function. SMARTLET (Self-organizing Modular Autonomous meso-Robotic Transistor-circuitry with Locomotion, Energy, Telecommunication and Sensors) is the generic acronym we give to denote the general class of modules with the full functionality to support microelectronic morphogenesis.

In fact, one can characterize current progress in functional technology with three axes as shown in **Figure 1**. The two spatial axes capture both the tradeoff between miniaturization and information processing capabilities. The diagonal corresponds to a constant limiting information processing density limit. The third color-coded axis, reflecting structural flexibility, is important for the use of information processing to perform shape changes and morphogenesis, which not only facilitates self-reproduction but also a variety of mechanical functions. While these three axes are perhaps not the only choice, they do capture minimal requirements for technology to approach life and show what needs to be achieved for this.

In what follows, we focus on the potential of modular microelectronic systems, whose modules like cells exhibit shape morphing properties that support the core functions above. This perspective driven review is inspired by the observation that we are now at the threshold to a new quality of technology that can combine high degrees of electronic information control with small dimensions and functionally significant shape changes. Its timeliness is confirmed by the appearance during its revision of an article implementing 3D folding control *via* single crystalline silicon microribbons[33].

At larger scales (> 1 cm), modular robots have been developed consisting of electronic functional modules[34] that form into complex distributed systems, inspired by the distributed functionality and self-organizing properties of multicellular organisms. Such modular robotic systems can self-assemble and reconfigure into various 3D shapes and self-adapt to an unknown and dynamically changing environment[35], the modules providing telecommunication, sensing, actuation, processing, energy storage and power management functions[34]. Modular rigid and soft robotic systems have become a vital platform to study self-organization and evolution, relying on self-assembly from a set of autonomous[36] kinematically free modules with embodied functions and interaction rules[37–39] that can connect or disconnect from each other, adapting the body shape. Collective modular actuation and locomotion of these macroscopically modular robots may lead to fascinating application scenarios in areas such as inspection, rescue, exploration, and navigation**.** However, the macroscopic size of their modules strongly limits their overall adaptability, mass construction economy, evolvability, as well as their self-sufficiency and sustainability.

At microscales (< 1 mm), current hardware implementations of autonomous microsystems are progressing fast but are still in their infancy when it comes to mimicking both the elementary and collective behavior of their natural counterparts. The two most prominent examples are smart dust[40] as a distributed surveillance system for body or environmental networks and miniaturized robotics[41]. Recent algorithms and technology[42] are opening new channels of low power communication that are enablers for electronic communication between smart modules, even in wet environments where electromagnetic communication is strongly damped[43]. The feasibility of powerful miniaturized robots was first convincingly argued in Richard Feynman's famous lecture series "There is plenty of room at the bottom (1959)"[44]. And what seemed physically feasible in principle, but remained unachieved for decades, now starts to take shape, thanks to the rapid development of materials science, microelectronics, nanotechnologies, and system integration[41]. The progress in microelectronics over the past few years has brought to life highly integrated remotely powered microscopic robotic systems with onboard digital control and integrated actuators.[45,46] Micro-robots have begun to swim, walk and carry out useful tasks both *in-vivo* and in the environment[47,48]. Spanning 3D space like voxels[49], these modules can dynamically self-evolve, accomplishing defined tasks while adapting to environmental changes. They have already been made of intelligent materials but, so far, they are not self-aware or self-repairing and their control requires significant computational resources.

But are current autonomous microsystems sufficiently advanced to tackle the three core properties of cells introduced above? Firstly, for homeostatic *self-maintenance*, energy harvesting and storage[50] as well as material self-assembly are required. They have both been demonstrated for microsystems, at least with highly structured energy (for example at specific wavelengths in the form of electromagnetic radiation, as solar energy or laser irradiation, or ultrasound acoustic waves, providing more energy than background noise[51]) or building materials (such as complex patterned blocks like magnetic voxels[52,53][49]). A complete synthetic metabolism – constructing the wealth of specific molecules required to build a cell having its own construction machinery – remains a formidable feat of biological cells, unmatched by chemical technology. While the origin of life may require organisms to be completely autotrophic, many cells are not: Mammalian cells for example, can only synthesize half of the twenty amino acids they need and require also at least two fatty acids as well as vitamins[54] (all stemming from other organisms). Operation in highly structured environments (with complex fabricated resources) is both legitimate and safe in the first steps towards material rather than simulated artificial organisms, simplifying the metabolic task, in a similar fashion to the supply of very specific monomers in synthetic biology systems, and removing any risk of proliferation in the natural environment. However, using a set of fixed building blocks, significantly more complex than amino acids, will limit the flexibility and evolvability of the system. In Section 5 we show how this limitation can be overcome, so that the organism can direct the synthesis of its own custom building-blocks.

For *self-containment* and self-identification, as achieved by membranes and specific molecular receptors for biological life – allowing a distinction between components that belong and do not belong to a cell by means of either being unambiguously on the inside or having the identifying code/substructures – we must distinguish two scenarios: the connected-component machine scenario, familiar from our everyday devices, and the organismic architecture, composed of many freely mobile components (*e.g.*, the RNA and proteins in cells). For the latter, the management of a boundary is important, allowing the passage of nutrients and waste but preventing the crossing of informational entities, that may from the outside be acting in the reproductive program of another organism or from the inside be vital to retain for the organism itself. The set (if non-empty) of informational entities in the boundary itself is accessible to other organisms and hence the outer shell or bounding container of the organism must as in the former scenario be able to distinguish informational components that belong to self from non-self. For this reason, although possibly not primordial, we see the distinction of those smart components belonging to the organism as a fundamental capability. Smart components with sufficient on-board electronic memory to include a hard-to-guess organismic identifier can complete initial identity checks on new components with which they dock in a secure initial handshake (like a lock and key) that rejects docking completion on failure. This process can also be utilized to enable error correction in the self-assembly process, allowing sophisticated and reliable component positioning. It requires the ability in proximity to exchange electronic information and the ability to actively accept or reject the docking of components.

*Self-reproduction* in non-biological technology appears to be both difficult, especially for three dimensional machines, and of dubious value, because of the concern about control of proliferation, despite early demonstrations of the value of technical evolution[18], and although it does find extensive use in molecular biotechnology, cellular biotechnology, and tissue engineering. Fortunately, nature has shown a path to achieving the replication of folded 3D structures that, if abstracted sufficiently, points the way to a resolution of both these difficulties and concerns. Nature's solution is to achieve replication of non-directly copyable complex structures (such as proteins or cells) by packaging them together with a replicable structure (such as DNA) and then encoding in the latter a recipe (list of fabrication instructions) for a set of chemical machines to build the former. In the case of proteins, the machinery involves the spliceosome and ribosome as well as many other proteins, RNAs as well as small metabolites and ions. In the cell, the machinery is a combination of the environment and cellular machinery like the mitotic spindle and molecular motors. In artificial organisms, we do not need to, and should not even try to get the organism to fabricate the entire construction machinery for its smart components. Instead, to enable full self-reproduction, within the capabilities of the microfabrication facility, it is sufficient for the organism only to specify the construction of some of the necessary self-assembling components, provided that the organism does carry a fabrication recipe (encoding the individual organism and its components) to operate the entire construction machine. This technique of indirect self-replication was employed in drug development on gel beads by Affymetrix, following ideas of Lerner and Brenner[55]. Thus, self-reproduction can be achieved by allowing an organism's smart modules to specify a recipe for their own fabrication. This recipe can be compressed (hash coded) succinctly, so that the encoding string length need only encompass a library of possibilities commensurate in size with the number of different recipes that will be constructed.

As seen in Figure 1, we propose that SMARTLETs are the general class of modules with the functionality to support microelectronic life. In this review, we will explore the progress towards microelectronic morphogenesis as a basis for electronic living technology deploying these smart mesoscopic (1-1000μm) modules that can actively control their shape and interactions. Self-assembling smartlets also addresses key additional functions of cells listed above such as shape changes and movement, environmental sensing, communication, and collective action. Shape changing (by bending actuators[56]) and movement functions (for example by micromotors [41,57]) have both been achieved with this microtechnology as we shall see below. Additional avenues of communication including ionic and chemical signals, ultrasound and electromagnetic waves may be employed. Chemical sensing, through the increasingly diverse arsenal of electrochemical sensors, provides a natural path to sensing the chemical environment (as well as internal chemical state), but there is an increasingly diverse range of other sensors – including magnetic sensors[58,59] or those related to internal stress[56] – that have been realized with fold-up thin film technology. Coupled with shape changes and locomotion these sensors allow the artificial organism to mechanically sense its environment. As we shall see below, collective motion of self-assembled smartlets can be organized through the electrical connections and integrated circuitry of the smartlets as can communication.

To develop this new potential systematically in this review, we first review the folding of smart thin film microelectronic modules and the achieved functionalities in Sections 2 and 3. Then we assess the progress being made in the self-assembly and interconnection of (folded) microelectronic modules in Section 4. Section 5 then reviews how these developments are enabling a new generation of modular microrobots. We then return to the question of how far this technology addresses the core functions of cells and organisms in Section 6. In Section 7 we take the perspective for morphogenesis of organisms one step further in connection with the differentiation and spatial deployment of different modules. We conclude with an outlook on the perspectives for radical progress with sustainability by embracing microelectronic morphogenesis to create a new living technology platform.

## 2. Folding of smart thin film microelectronic modules – shape diversity

The folding of linearly synthesized structures like proteins to form enzymes is a key enabling technology for life. The folding of modular 2D patterns (often in the form of branched linear chains of modules like the cruciform structure that folds to a cube) on thin films to 3D structures is poised to play a similarly vital role in the creation of functional microelectronic modules (**Figure 2**). The conversion of linear genetic sequence information to function in biological cells involves the folding of biopolymers like RNA and proteins to 3D structures (Figure 2A). This is central in biology because the linear structures (templates) of DNA and RNA are both directly replicable and can be folded to the non-replicable three-dimensional forms that allow complex functions. Structure formation plays a dominant role as a determinant of biological function. Indeed, one can search for catalysts for reactions by requiring them to bind to the transition state structure[60]. 2D planar patterns generated by photolithographic masks, whether directly written or copied from a master, continue to be the mainstream of high-density electronics. To maintain Moore's Law, multiple layers of transistors in the same silicon crystalline structure or more recently vertical stacks of thinned chips with high-density vertical interconnect or interposers[61] have been produced as layered 3D structured devices. Contrast this with the ability of planar structures to fold into non-space-filling three dimensional architectures[62]. In Figure 2B, we exemplify this "2D genetic" encoding of the folding of planar sheets into secondary and tertiary structure with either gradual (e.g. roll-up) or sharp (cuboid or more generally origami- or kirigami-like) folds. The primary information in the multilayer 2D patterned stack, which may, as in kirigami, be cut to involve only a fraction of the plane, results in specific secondary structures which can further self-assemble in 3D to specific complex functional modules.

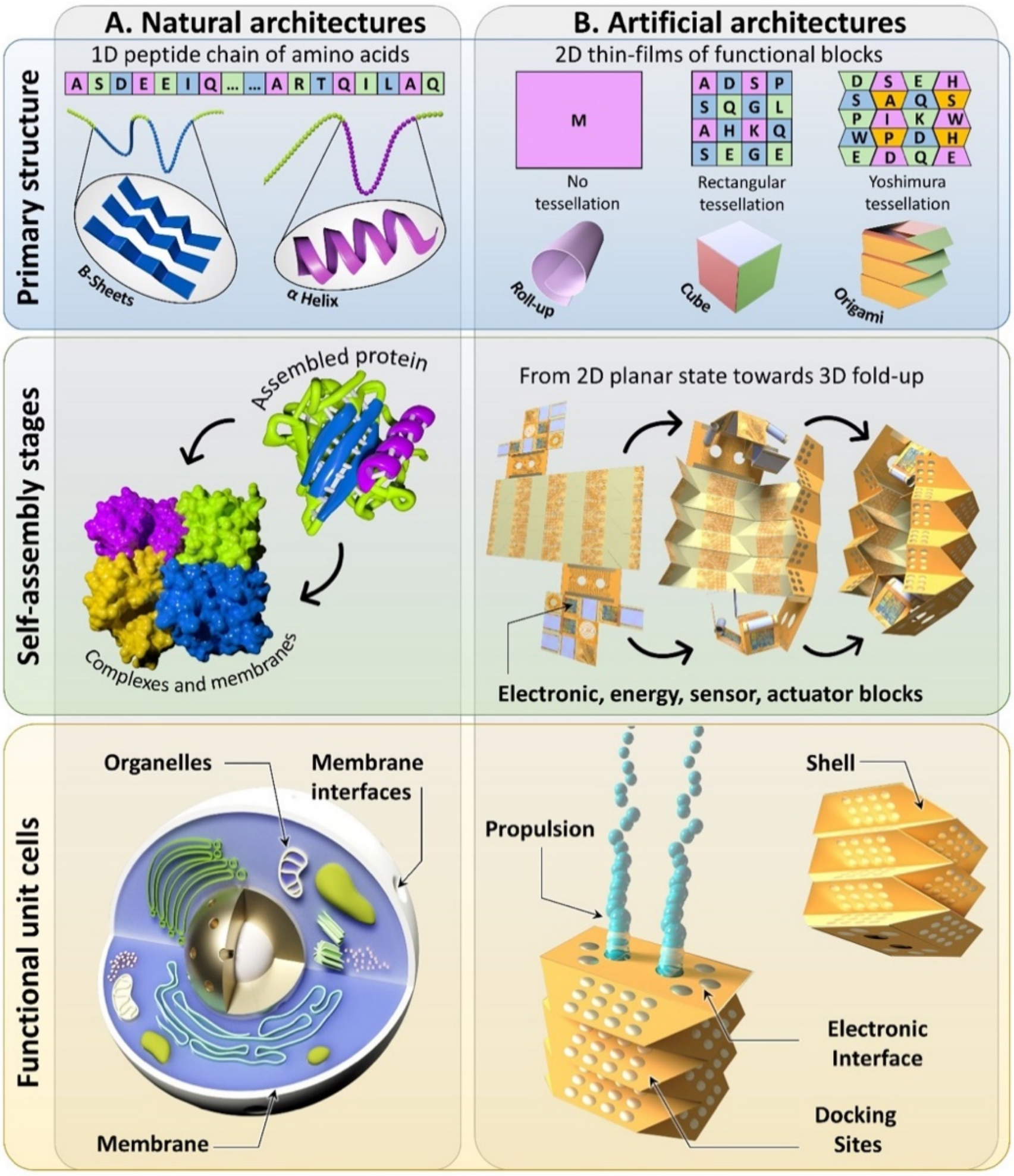

**Figure 2: Architectural encoding of cell units for natural and artificial organisms.** The two columns illustrate the remarkable analogy between natural and artificial architectures that are assembled from 1D linear chains and planar 2D-structured thin-films respectively. **A.** 1D genetic encoding of biopolymer sequences which fold into 3D structures like enzymes self-assembling into natural cells **B.** 2D encoding of multilayer pattern elements using microsystem technology, which fold up into 3D structures, self-assembling to form microelectronic SMARTLETs.

The conceptual advantages of this 3D self-assembly approach are simplicity and access to complex architectures that are often infeasible when using traditional techniques and are foreseen to be ultimately hierarchical, scalable, and parallel. The integration of microelectronics into the self-assembly of modular 3D structures is illustrated in **Figure 3** and animated in **Video S1**. These structures can be predicted algorithmically[63] providing a specific yield of a certain self-assembly pathway. Approaches for wafer scale fabrication of 3D architectures have been recently reviewed by Kwok et al.[64] focusing primarily on surface tension driving forces, by Lazarus et al.[65] with the focus on strain generated in metallic thin films, and by Karnaushenko et al.[66] with the focus on shapeable material technologies offering polygonal (see Figure 3 for design) and tubular "Swiss-roll" architectures that are compatible with microfabricated flexible thin-film and rigid silicon electronics.[67] Tubular "Swiss-roll" architectures (see also Section 3) offer high intrinsic surface area and allow the combination of cylindrical and helical geometries while the polygonal architectures provide flat surfaces for strain critical functionalities such as magnetic sensors[68] and silicon chips. Microelectronic precision and wafer-scale manufacturing of these self-assembling architectures demonstrated high-fabrication yields[69][70] (both structural and functional >90%) and throughput (>1000 of structures per 6" wafer).

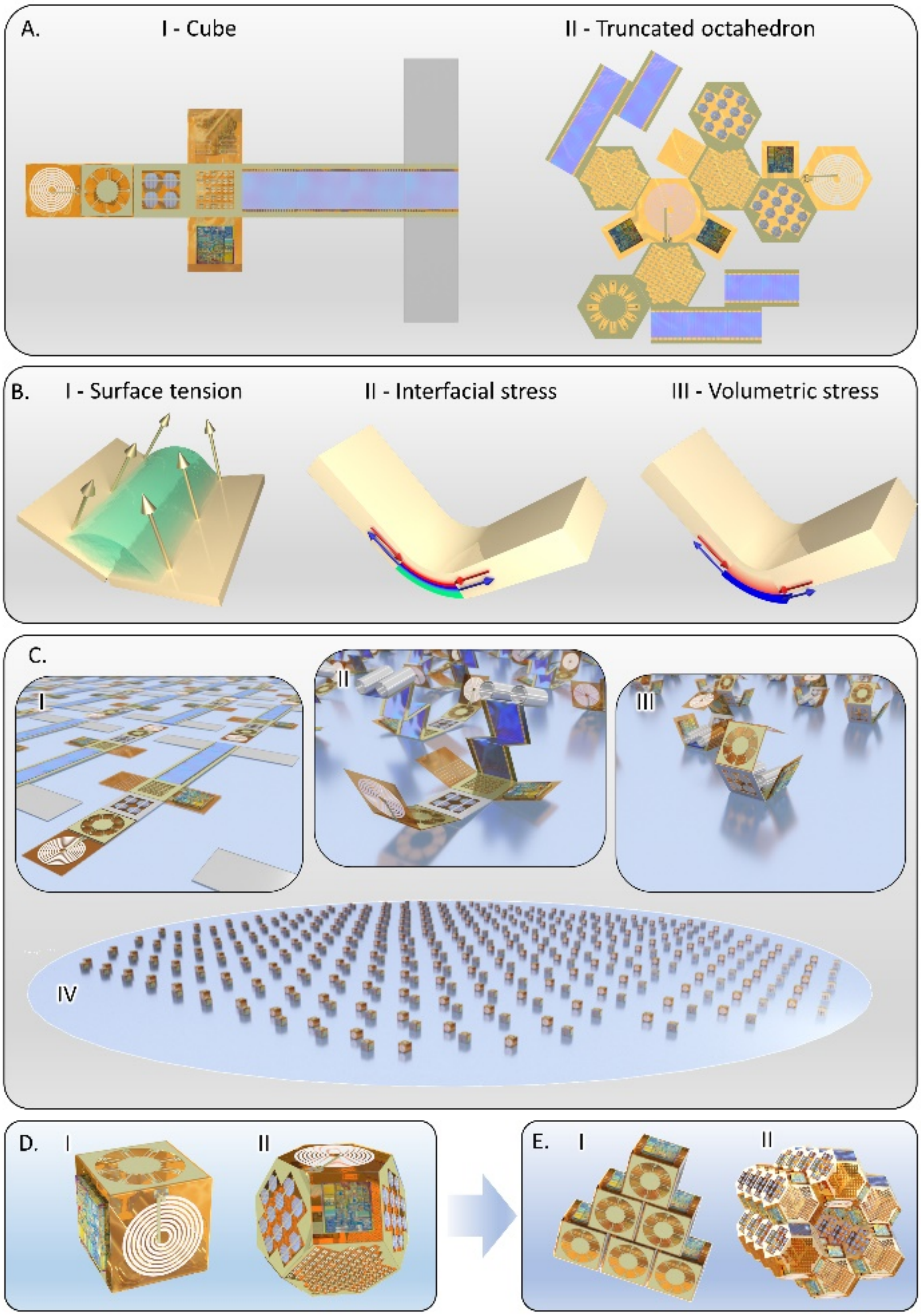

**Figure 3: Microelectronic pathway for morphogenesis. A.** Designs of planar layouts that integrate a variety of electronic functions and can re-shape themselves into 3D structures: a cube and a truncated octahedron respectively. **B.** The fold-up self-assembly of these structures is driven by commonly used physical forces like I. surface tension, II. stress at the interface of thin-films or III. volumetric expansion of materials like hydrogels. **C.** Self-assembly occurs in parallel for all the planar structures fabricated on a wafer: I-III successive folding stages closeups, IV completed folding on wafer. **D.** Self-assembled architectures equipped with micro-electronic functions form SMARTLETs (basic active building blocks): see Video S1 for animation. **E.** SMARTLETs can then be aggregated passively or actively into higher hierarchical assemblies: I – cubes; II – truncated octahedrons.

There is a remaining difficulty in the reproduction rather than mass-production of such folded structures: given a 3D structure one needs to extract the information about primary structure in order to copy it by planar fabrication and folding. For SMARTLETs, like proteins, accessing and reading out physically the lower dimensional material patterns used to make the 3D structure is not generally achievable. However, the 2D masks and fabrication protocol used to make the structure can be digitally encoded in the electronic memory of a CMOS chiplet integrated in the SMARTLET and readout wirelessly. It is not necessary to restrict the set of spatial motifs and fabrication variants used to make smartlets to a small number (like the 20 amino acids of proteins) for this, because hash coding can be used to create a fab code which allows a complex fabrication protocol to be uniquely encoded (*e.g.,* with 64 or 128 bits of information). Machines that can read out this fab code can feed the information to a clean room fabrication line to replicate the building blocks. In addition, the deployment protocol required for self-assembly of the voxels to a particular organism can likewise also be encoded in them (also enabling self-identity control algorithms by the organism). The deployment environment could be either serial (delivering voxels one by one in a prescribed order for self-assembly) or parallel (specifying the complete set of voxels needed) or any intermediate between these two. Note that motorized smartlet voxels can also navigate to a predetermined or computed sequence of locations to accelerate and improve the accuracy of the self-assembly process.

## 3. Functionalities of folding thin film microelectronics

As recently reviewed [71], thin film active electronics with extended functionality beyond electronic information processing has been the subject of major development in the last few years [72,73]. Extended functionality includes actuators[56] and sensors[56,74], microbatteries[75] and the integration extends all the way to microrobots[41]. Recent breakthrough achievements enabled by fold-up electronics include 3D electronic and sensor voxels[76–78], microbatteries[75], energy harvesters[79], bending actuators[80] and microengines[57,81] and we have collected key developments integrating electronics into 3D microstructures in **Figure 4**.

The conceptual advantages of this 3D self-assembly approach include simplicity and access to complex architectures that are often infeasible when using traditional fabrication techniques. They are also ultimately hierarchical, scalable, and parallel, and deliver folded architectures (for example polygons and tubular "Swiss-rolls" as shown in Figure 4A) that are compatible with microfabricated flexible thin-film and rigid silicon electronics[67]. Examples of 3D electronic structures making use of these architectures are shown in Figure 4B. Rolled ("Swiss roll") and folded architectures increase energy density by filling surface patterned active material into a given volume. The compact form factor optimizes space utilization, suitable for acting as power units at the microscale. The extension to dynamic (4D) structures is well underway as shown in Figure 4C. The ability to make and break high quality (sub-ohmic) electrical contacts[82] between modules by self-assembly (Figure 4C-V) is a key development, allowing complex circuitry to be generated and tested during growth. Folded hybrid architectures[83] such as microfluidic containers[84] and channels with electronic walls open possibilities to store and transport chemicals and materials in artificial cells and through organisms even against their thermodynamic equilibrium. The manufacturing and 3D self-assembly of layered planar thin-film structures relies on conventional microfabrication processes and new functional lithographically processable materials that exhibit electrical conductivity[85], stimuli responsiveness[86], and chemical[87] and optical functionality[88].

In addition to these architectures, tessellated origami folding architectures (Figure 2B) offer extreme mechanical flexibility[89] in some directions, as well as structural rigidity in others, allowing both reconfiguration and actuation[90] of the assembled module parts. Folding architectures also provide a large surface area for integrating electronics, energy, and other functions, which are needed to construct self-sufficient dynamic smartlet modules. Current achievements are still only early demonstrations of the powerful capabilities provided by this novel integration scheme which brings together microfabrication, conventional and thin-film microelectronics, 4D architectures and functional soft materials.

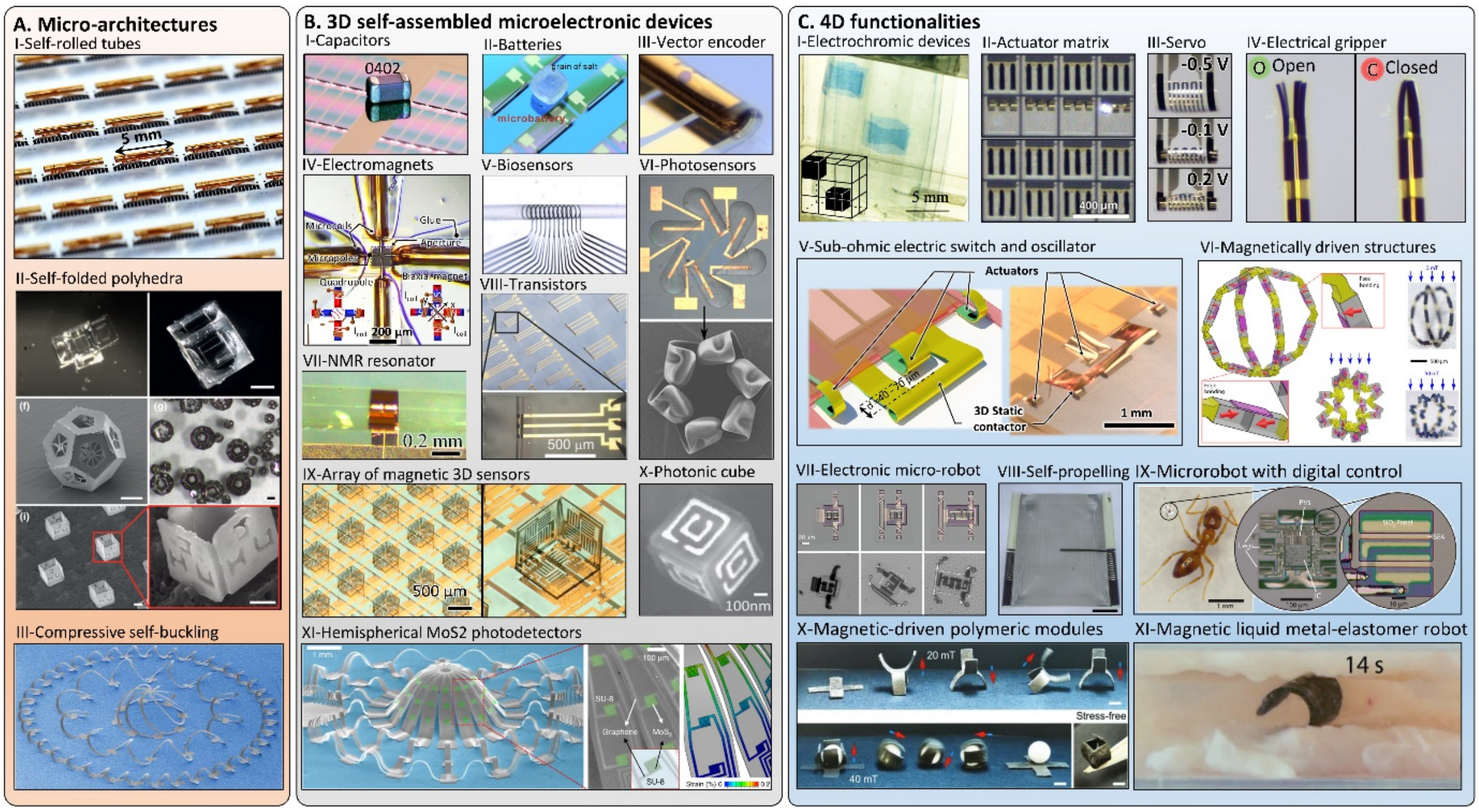

**Figure 4: An overview of self-assembled microarchitectures, 3D functional electronic devices and time dependent functionalities. A.** Microarchitectures that can be formed by e.g. I - self-rolling ([59] † © 2019 TA, PB American Association for the Advancement of Science AAAS), II - folding planar sheets (AP [91] © 2011 Elsevier Ltd.), or III - controlled compressive buckling (AP[92] © 2015 AAAS). **B.** 3D self-assembled microelectronic devices that rely on these 3D architectures: I – electrostatic capacitors ([69] †© 2020 TA PB W); II – batteries ([50] †© 2022 TA PB W); III – Magnetic 3D vector field encoders ([59] †© 2019 AAAS); IV – Microelectromagnets ([70] †© 2022 TA PB Springer); V – microfluidic systems and biosensors ([93] ‡© 2021 TA PB W); VI – graphene photosensors ([94] †© 2020 TA PB W); IX – active matrix 3D magnetic field sensor array ([68] †© 2022 TA PB Springer); X – THz photonic architectures. AP[95] © 2011 W); XI – optical detector array. ([96] †© 2018 TA PB Springer). **C** 4D functionalities: I – Assembled voxelated electrochromic display. (AP[97] © 2007, IEEE); II – Active actuator matrix ([93] †© 2021 TA PB W); III – feedback-controlled actuators ([98] †© 2021 TA PB W); IV – electrically driven grippers ([99] †© 2021 TA PB AAAS); V – sub-ohmic electric switches and mechanical oscillators ([82] †© 2021 TA PB AAAS); VI – Magnetic shape changing structures ([49] †© 2021 TA PB AAAS); VII – Electronic microrobots (AP[100] © 2020 Springer); VIII – Self-propelled microelectronic system (AP [101] © 2020 TA under exclusive license to Springer Nature Limited) IX – Microrobots with integrated digital control (AP[46]© 2022 AAAS); X – Magnetically driven modules (AP[102] © 2021 W); XI – Magnetic liquid metal elastomer robots ([103] †© 2022 TA PB W).
† Adapted under the terms of CC BY-NC 4.0 license. ‡ Adapted under the terms of CC BY-NC-ND 4.0. W WILEY-VCH Verlag GmbH. AP Adapted with permissions. TA The Authors. PB Published by

We complete this section with a review of current progress in integrating power units such as microbatteries into functional modules at this scale (**Figure 5**). The principal advantage of integrable energy storage lies in its capacity to transcend spatial and temporal limitations of power supply through remote powering via radio frequency (RF) and ultrasonic (US) waves, as well as energy harvesters such as photovoltaic (PV) cells. Consequently, this advancement expands the capabilities of smart assemblies. Within the realm of Swiss-roll architectures, layered structures for both batteries and supercapacitors have reached the sub-millimeter scale, surpassing even sub-0.1-mm$^2$ dimensions. The energy yield per unit area ranges from 0.3 to 1.5 milliwatt-hours for configurations with fewer than 10 windings, making them suitable for electric circuits and sensors with power consumption levels below 1 microwatt[104,105]. For low-power functions this may suffice, allowing simple integration of such units into functional modules (Figure 5A). However, to meet the requirements of locomotion and actuators necessitating frequent high peak power, further enhancements to energy density are imperative, and this can be achieved by incorporating additional folding or rolling iterations. The mobility of these power units envisions a charging process facilitated by docking stations equipped with remote power capabilities or energy harvesters, followed by the return to the assembly (Figure 5B). Another avenue for powering electronic functions within smart assemblies involves harnessing energy from fuels in the vicinity. Nevertheless, conventional fuel cells exhibit a preference for high temperatures to optimize their efficiency. Biofuel cells, in contrast, often grapple with low voltage and limited power output. The intricate nature of enhancing efficiency under mild conditions poses significant challenges in microdevice design. An alternative, promising approach entails harnessing energy directly from the environment through redox reactions catalyzed by appropriate catalysts (Figure 5C).

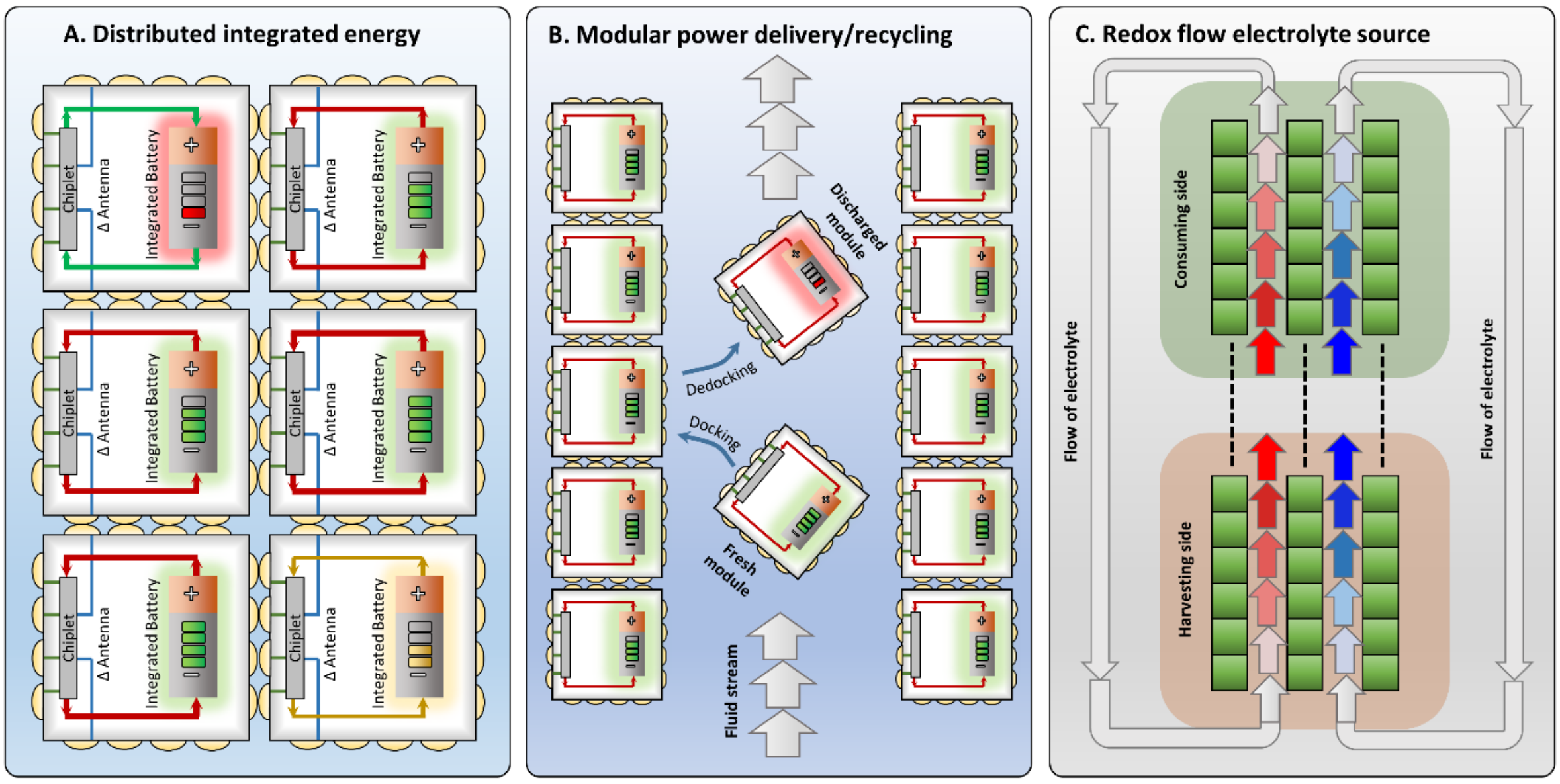

**Figure 5. Modular power units enabling the electric functionalities of smart assemblies. A.** Distributed integrated energy approach employing integrated batteries on independent modules to power low-power functions. **B.** Modular power units that can be obtained from the charging stock and assembled into a power pack to support high-power functions, with the ability to be recycled for recharging. **C.** Integrable power units utilizing redox flow electrolyte for continuous operation of electric functions. The electrolyte flow could be transported in self-assembled microfluidic networks

## 4. Self-assembly and interconnection of (folded) microelectronic modules: Operation during construction enables homeostasis

Whereas single biological cells, with their sophisticated molecular metabolism, maintain their material form and function homeostatically, microelectronic organisms exhibit homeostasis at the granularity of SMARTLET modules, when many of these self-assemble into more complex structures, through the dynamic exchange of modules maintaining the organism's structure. Automatic self-maintenance with building blocks requires forming a stable dissipative structure. This involves concentration of blocks (for example, by employing microfluidic concentration[106] into a reservoir), self-assembly and disassociation kinetics that scale differently with organism size. However, electronic component self-maintenance can be considerably more sophisticated, making use of analog (e.g., completion of resonant structures) or digital (counting required number of components) properties to provide signals to limit and maintain the growing electrical circuit. In this section, we review progress towards modular self-assembly of electronically active systems, addressing also the self-assembly of electrical and material (e.g. microfluidic channels) interconnects. **Figure 6** presents an overview of current progress towards the self-assembly of SMARTLET modules. This includes 3D electronic modular systems in Figure 6A, modular microfluidics in Figure 6B, modular microrobotics in Figure 6C which can be combined with the magnetically directed self-assembly of Figure 6D-G. The proliferation of electronic components necessitates power units capable of sustaining the escalating energy demands of the expanding electrical circuits. A prevailing approach to meet the power requisites of devices across diverse scales entails the utilization of modular batteries, configured into assemblies. Mobile tubular microbatteries exhibit potential as modular units that can be recharged through remote charging docks or by harnessing redox species available in the environment, such as the dissolved oxygen present in fluids[107,108]. Of special note is the high-density control of docking achievable with tubular docking in Figure 6E. It is important to realize that interconnect at the level of module-to-module docking is not the only mechanism enabling directed transport of power, information, or materials. For example, channels formed by the self-assembly of modules can form artery-like channels as highways for the transport of modules and other materials as shown in Figure 6H.

One of the paradigm changes proposed in this work is that instead of building electronic devices which are then turned on after completion and either function or do not, with no way of correcting erroneous function or repairing subsequent damage to themselves, active electronic circuitry can be included in the modules which is sensitive to the completion of correct assembly. If the modules' docking properties can be modulated by the electronics, then active local disassembly and reassembly can be triggered by this sensitivity, enabling homeostatic control of complex structures. Furthermore, it has been shown that self-assembly of simple circuit modules can enable the evolution of complex two-dimensional circuits such as digital multipliers effectively.

The self-assembly and self-repair of artificial organisms can potentially be made vastly more efficient and reliable through communication between unconnected SMARTLETs. There are of course other application advantages of communication, but the dynamic control of the organism's structure is a fundamental target for communication worthy of first attention, forming a platform capability equipping the organism for other distributed tasks. Self-propelled smartlets able to communicate during self-assembly can co-ordinate their motions and docking collectively to optimize the efficiency of self-assembly to be orders of magnitude faster than for random motion. While locomotion may not be sufficiently accurate for piloting docking, it can bring smartlets into proximity to allow local pattern matching potentials (*e.g.*, magnetic or shape encoded or both) to complete (or reject) docking.

While smartlets in direct electrical contact acquire efficient communication through the self-assembled electrical circuit, signal transmission through the surrounding media is needed for non-contact communication. RF (radio frequency) communication was the signal transmission proposed for smart dust[40] and then for the Internet of Things[109] (IoT), but it is not always the best choice. For small systems, RF reception and transmission are limited by the size of the antennae deployed and become increasingly inefficient for mm to µm sizes, even for > 5 GHz bands. Optical peer to peer communication has been achieved for microparticles at near 100 µm scale[110], but while reception with photodiodes is now highly integrated and efficient, as in the 1 μm pixels on smartphones, the power required for transmission using LEDs is still quite large and light is also reflected by many smartlet materials and media, limiting communication to line of sight.

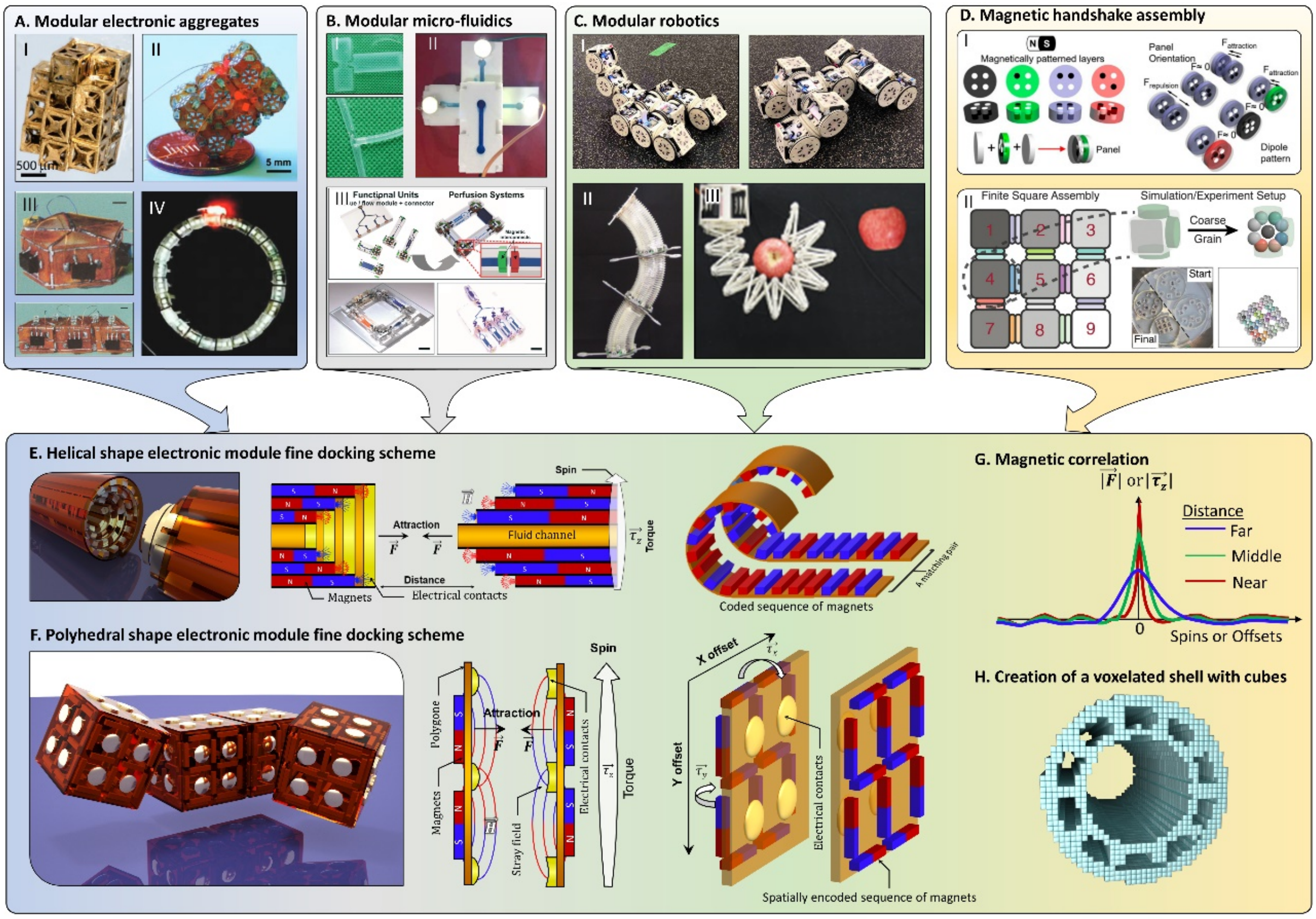

**Figure 6: Modular systems that explore surface tension, magnetic field and mechanical docking for self-aggregation. A.** 3D electronic systems: I – metallic cubes (AP [111] © 2010 American Chemical Society); II - electronic circuit (AP [112] © 2000 AAAS); III – distributed logics (Adapted from[113], © 2002 The National Academy of Sciences, USA); IV – electronic circuits with mobile magnetic modules ([114] †© 2021 TA PB W). **B.** Modular microfluidic: I – T-junction self-assembled modules ([115] †© 2022 TA PB W); II – Microfluidic blocks with capillary force alignment (AP [116] © 2018 IOP Publishing Ltd); III - Magnetically interconnected microfluidics[117] ([49] †© 2021 TA PB AAAS). **C.** Modular robotics: I – SMORES-EP modular robot capable for parallel self-assembly (Adapted under the terms of exclusive license to Springer Science + Business Media, LLC.[118] © 2023 TA PB Springer); II – Origami continuum robot (AP [119] © 2021, Mary Ann Liebert, Inc.); III – reconfigurable continuum robot based on tensegrity building blocks ([120] †© 2023 TA Advanced Intelligent Systems PB W). **D.** Magnetically encoded assembly: I - scale-invariant magnetic platform for programmed self-assembly ([121] ‡© 2019 TA PB PNAS); II – encoded hierarchical magnetic self-assembly (AP [52] © 2022 The Royal Society of Chemistry). **E.** Docking scheme for cylindrical modules that includes magnetic encoding, electrical sub-ohmic connection and fluidic channel. **F.** Docking scheme with polyhedral (cube) shape modules magnetic encoding and electrical connection. **G.** Magnetic forces and torques generated by the integrated micromagnets guide modules during docking towards the correlation maximum of complementary magnetic spatial structures. **H.** An example of a 3D shell structure made out of cube modules forming a voxelated channel for transport, *e.g.* fluidics.
† Adapted under the terms of CC BY-NC 4.0 license. ‡ Adapted under the terms of CC BY-NC-ND 4.0. W WILEY-VCH Verlag GmbH. AP Adapted with permissions. TA The Authors. PB Published by

Ultrasound at MHz frequencies is an efficient underwater communicator with greater penetration depth in the body[122] but electrical detectors whether piezo (PMUTs)[123,124] or capacitive (CMUTs)[124,125] are currently 100x larger than for optical systems and need larger voltages, complicating fabrication. Ionic potential transmission has recently been proposed for intrabody communication between implants[126] and may prove to be a good solution for smartlets in aqueous solution. All these high frequency communication channels depend on onboard integrated microelectronics and are not available to natural organisms, so that we may expect fundamental advances in construction and dynamic reconstruction (morphing) capabilities.

Since the main energy use in all these communication methods is for transmission, it is advantageous, at least in the initial phases of artificial organism development, to consider indirect communication through an externally powered network that can track the responses of the smartlets (*e.g.*, back scatter of RF as in RFID tags) and dynamically program their active motion individually or according to a classification of their IDs. Later, depending on progress with integration and power management, this can be extended to a single organism-based transmitter and ultimately to all smartlets carrying transmitters. Centralized control requires tracking or imaging technologies that are capable of 3D reconstruction and module tracking in real time.

## 5. How this technology addresses the core functions of cells and organisms

Considering this technological progress, it is time to return to the question of how far this technology can take us towards the core functions of cells and organisms – recalling that we replace the currently unrivalled compactness in cells of evolved metabolic chemistry for biopolymer monomer synthesis by fine-grained information access to advanced clean room fabrication of the 2D templates for SMARTLETs. Thus, while a microelectronic organism cannot fabricate new or replacement SMARTLET components on site only from environmental resources, it can (ultimately autonomously) send an order in the form of coded fabrication instructions (initially as simple as a choice between a finite number of alternative forms) to the fab line for delivery of specific modules. Note that just as cells instruct the synthesis of complex arrangements of modules in RNA and proteins by copying the extant neighbor relationships between monomers, it may be desirable to synthesize connected or spatially arranged assemblies of modules rather than relying completely on free-space self-assembly, extending the folding approach of Figure 2. In this section we first consider the efficacy of self-assembly in the context of possible enhancement systems more systematically, before turning to the ability to scale down current achievements further towards the scale of biological cells, to make fine-grained microelectronic organisms.

Although the sorting into a 3D structure of randomly positioned elements only by self-assembly may not scale well with the number of components for reliable single-level (as opposed to hierarchical) self-assembly, especially for microscale components where the frequency of component-component interactions is limited compared with molecular scales, there is already a repertoire of ideas for overcoming this limitation. Six main approaches to assisted self-assembly by pre-arranging modules in space can be pursued: (i) *addressable self-assembly*[127] – retaining at least 1D physical connection between modules during 3D deployment (as is done with biopolymers). Net or branched chain connection of structures makes use of kirigami-like folding of cut or lithographically separated shapes and has even been optimized shapes like platonic solids[128]. (ii) *layered proximity* – retaining 2D spatial planar ordering of modules to a significant extent upon release for self-assembly (thus partially guiding self-assembly in layer-by-layer release, as modules then only need to dock to the correct one of the few modules in the local vicinity). Although layer by layer self-assembly has been employed for polyelectrolytes[129], to the best of our knowledge this approach has not yet been employed for released module self-assembly. (iii) *serial module printing* – 3D printing modules to the correct approximate location for the assembled structure (relying on local self-assembly for final fine positioning and docking). Serial module printing *e.g.,* based on jamming microgels, has been employed primarily for biological cell printing[130,131] but printing of flexible electronic structures with these techniques has been envisaged. (iv) *selective transfer printing* in parallel from arrays of components using a variety of techniques such as epitaxial or laser lift off[132] and soft stamp micro transfer printing[133] (v) *differentiation and cell division* – which is the natural route for multicellular organisms. The differentiated progeny of a given cell-type are formed by cell division and differentiation at specific locations (often determined by diffusing morphogens or specific patterns of mechanical stress[134]). Differentiation will be discussed further in Section 6. (vi) *Locomoted self-assembly* – in which the modules contain information pertaining to at least the approximate 3D location (whether in relative or absolute terms) or their target location and have a means for locomoting to this location. Combinations of these methods with full self-assembly (unsorted release of modules) and with one another are also potentially useful.

How fine can SMARTLET modules be constructed? Can they reach a similar scale to cells (*e.g.,* 10 µm). The physical limits of nanomorphic cells (with biochemical functions replaced by bioelectronics) were discussed over a decade ago[51], with an affirmative answer to their viability in principle at cell size, but without a clear pathway to effective realization. Of course, microelectronic morphogenesis based on folding of smart thin film structures with bonded chiplets (SMARTLETs) is already interesting at any sub-mm scale, but collective structures become finer and potentially more intricate as the size is reduced.

**Table 1. Projection of scaling of 4-bit microcontroller capacity CMOS chiplets down to cell sizes.** The technology node years for the last three steps are adapted from the largest foundry (TSMC). Reference object and chiplet sizes are shown on the left. Data is adapted from http://www.europractice-ic.com/.

| Reference size object | Size µm | Techn-ology nm | Transistor number[d] | Voltage core/IO volts V | Memory (SRAM) bits | Pad sizes µm | Supercap capacity nF | Load[b] pW |
|---|---|---|---|---|---|---|---|---|
| lablets [c] | 140 | 180 | 2238 | 1.8/3.3 | 843 | 28 | 4900 | 2000 |
| | 140 | 130 | 4290 | 1.2/2.5 | 1613 | 28 | 4900 | 2000 |
| | 100 | 90 | 4360 | 1.2/2.5 | 2002 | 20 | 2500 | 1000 |
| human hair | 70 | 65 | 4185 | 1.2/2.5 | 1867 | 14 | 1225 | 500 |
| | 45 | 40 | 4050 | 1.1/2.5 | 1674 | 9 | 506 | 200 |
| | 32 | 28 | 4096 | 1.0/1.8 | 1575 | 7 | 256 | 100 |
| neurons | 24 | 22 | 3732 | 0.5-0.8/1.8[a] | 1435 | 5 | 144 | 60 |
| leukocytes | 12 | 10 | 4608 | V multiple | 1737 | 2.5 | 36 | 15 |
| erythrocytes | 6 | 5 | 3475 | V multiple | 1336 | 1.2 | 9 | 3.6 |
| bacteria | ≤3 | 3 | 3136 | V multiple | 1206 | 0.6 | 2.2 | 0.9 |

[a)] Using SOI (silicon on insulator) and local voltage back biassing, e.g. Global Foundries; [b)] With slow system clock in <kHz range or event driven; [c)] In 2016, 140x140x35 µm lablets were completed as tiny autonomous programmable electrochemical labs; [d)] 4-bit microcontrollers have been designed with *ca.* 4000 transistors (U. Tangen, T. Maeke, J.S. McCaskill, MICREAgents project, unpublished work) so this is taken as a reference number for programmable functions. Lablets with roughly half this number were also programmable but through a 58-bit shift register which modifies a finite state machine.

For SMARTLETs outlined above, there are several potential limits to miniaturization:

(i) Feature size for active folding elements. The feature size for active folding of thin films has been demonstrated with large angle bending over µm distances enabling near cell scale microrobots[135]. These actuators can be powered at low voltage (< 1V) and require immersion in an appropriate electrolyte, which in principle could be carried on board, although this has not yet been realized. This contrasts with many other actuator mechanisms (*e.g.,* microhydraulic, piezoelectric), where 30 - 100 µm seem more likely limits.

(ii) Device size for powering and communication. The device size for remote powering depends on the required power and powering mechanism. RF reception and transmission are limited by the size of the antennae deployed and inefficient for mm to µm sizes, even for > 5 GHz bands, strongly absorbed in aqueous media. Photovoltaic actuated mass fabricated robots can operate with only 10 nW of power, which can be extracted from incident sunlight on a 30x30 µm surface (100 nW) [136]. Optical peer-to-peer communication has been achieved for microparticles at near 100 µm scale[38], but the power required for transmission using LEDs needs further reduction and line of sight transmission must be complemented by other methods. An alternate powering via ultrasound (PMUT/CMUT) has the advantage of penetration depth within the body and has been proposed down to the 10 µm scale for smart neural dust, or body dust (for metabolic applications), although not yet realized below 100 µm[137]. Ionic potential transmission has been proposed for intrabody communication between implants[39] and may prove to be a good solution for SMARTLETs in aqueous media. Since external communication can be overlaid over the power channel, similar constraints apply to communication for these channels – at least for reception. However, as mentioned above, high frequency communication is enabled by onboard microelectronics not available to natural organisms.

(iii) Chiplet size for sufficient information processing. Chiplets with sizes of 100x200x35 $\mu m^3$[138] and 210x340x50 $\mu m^3$[139] have been realized with full CMOS functionality, using a 180 nm node process. A calculation of the scaling of logic resources with the size of chiplets is shown in **Table 1**, based on the resources required for programmable lablets or elementary 4-bit microcontrollers. Of course, the chiplets also need to be bonded effectively to the thin film structures and this also affects the achievable miniaturization. Chiplet bonding to rigid and flexible substrates has been achieved down to the scale of 20 µm and below[132,140].

An alternative approach to powering microelectronic morphogenesis is to employ a material energy currency (power modules), which can be fabricated as a special simple kind of disposable or recyclable SMARTLET. This becomes possible only because of recent developments in high quality (sub ohmic) electrical docking. The power can be in the form of a microbattery[75], as discussed above, with the folding technology scalable down to 20µm size, or involve chemical fuel. With these power modules structured as a SMARTLETs, with the same intelligent self-assembly, docking, and de-docking capabilities, but without the need for additional functionality to support the organism, their deployment can be directed to the target location by the same enhanced self-assembly options as discussed above. While this self-assembly of power modules has not yet been shown, it does not require any further new discoveries for technological realization. Because of extant microbattery achievements, this approach shifts the power delivery problem to more complex self-assembly and kinetic control rather than extracting power from a radiation field.

With respect to self-containment, firstly the electronic assignation and updating of individual electronic identifiers to SMARTLETs enables them to be used in electronically controlling docking during self-assembly to distinguish self from non-self at the module level during self-assembly and homeostasis. Such identifiers will already be valuable at 64 or 128-bit length and can be statically encoded in flash ROM at fabrication time and augmented with e.g., 32 bits

SRAM memory to record important deployment events or differentiation (see Section 6). The memory capacity of CMOS chiplets can accommodate this readily down to cellular dimensions (as shown in Figure 3). Secondly, SMARTLETs can also roll or fold up to contain chemical fluids without leakage as demonstrated already[141,142]. Thus, they can contain chemical resources and prevent access of external chemicals to their interior. It remains to be demonstrated that self-assembled SMARTLETs can make tight seals to retain fluids in folded sheets.

Finally, we outline how self-reproduction can be achieved in terms of the above SMARTLET technology, with the synchronization controlled by the active electronics of the self-assembled cells. As with multi-cellular organisms, the genetic description (describing the fabrication and deployment protocol) not only of individual SMARTLETs but of the whole offspring can potentially be encoded in individual SMARTLETs, at least in compressed form. This information needs to code for the fabrication and deployment (delivery, whether spatiotemporally sorted or unsorted) of a finite set (typically 1-10) of different SMARTLETs – using one or more of the mechanisms of self-assembly enhancement above. As electronic information in a SMARTLET, a single SMARTLET sent or extracted to the fabrication facility, or a wireless readout procedure can serve to deliver it from the organism to the fabrication and deployment machinery. The degree of autonomy versus filtered control in self-reproduction for organisms can thus be decided with full human intervention options. Self-reproduction at the organism level is thus an extension of self-repair mechanisms at the SMARTLET level.

## 6. Outlook towards microelectronic morphogenesis and differentiation

Building on the argued ability of microelectronic morphogenesis to integrate basic functionalities of cells and organisms, we now address what is available, what needs to be done, and how feasible it is, to generate differentiated artificial organisms with the properties of multicellular organisms. We will exemplify this with a target electronic organism self-assembled from SMARTLETs that has a similar cellular complexity and morphology to the eutelic biological organism, the rotifer, shown in **Figure 7**. An animation of the hierarchical assembly process and organism movement is presented in **Video S2**.

The rotifer, or "wheel" organism, is an aquatic microorganism which develops from a single cell to a differentiated multifunctional gastrulated organism with a species dependent constant number of cells (typically *ca.* 1000, *i.e.,* it is eutelic). It is capable of complex differentiated motion both of its "wheels" for feeding and of its body. The figure shows how SMARTLET modules can be folded up as structures with programmed flexibility that is retained during self-assembly via docking, while allowing electrical contacts to be made, giving the whole organism dynamic flexibility. The self-assembly at microscales would normally be slow, but the process can be radically accelerated compared with biological morphogenesis, by locomotion that is directed (driven) by communication (direct or indirectly *via* an external processor) between SMARTLETs with specific IDs. The self-assembly process, resulting in folded shells, also allows accessibility for the repair of the structure, replacing defective smartlets, as shown. **Video S3** shows how this modular repair can occur within the organism.

While cellular differentiation depends on genetic control through different genetically regulated states, the options for microelectronic morphogenesis open to electronically controlled SMARTLETs are either similar electronic induced functional differentiation of already assembled modules or the selective assembly of specific already differentiated SMARTLET modules. We address the former first. Just as differentiated cell states arise through the execution of different RNA expression patterns in cells from the same DNA, through a succession of developmental switches, nonlinear electronic state machines with operating memory can enter different and robust domains of operation to control smartlet cell operation. The posited ability of differentiated SMARTLETs to communicate significant portions of their chiplet memory (their somatically differentiated electronic genome) to other proximate "bare" SMARTLETs, which have self-assembled in their vicinity in their undifferentiated foundry state, allows them to effectively self-reproduce, creating differentiated SMARTLET tissue. A near-microscopic chiplet (at 100 µm scale, called lablets because of their functioning as an electrochemical/electrokinetic laboratory) able to send its program to another similar chiplet was designed and constructed as part of the MICREAgents project[143,144].

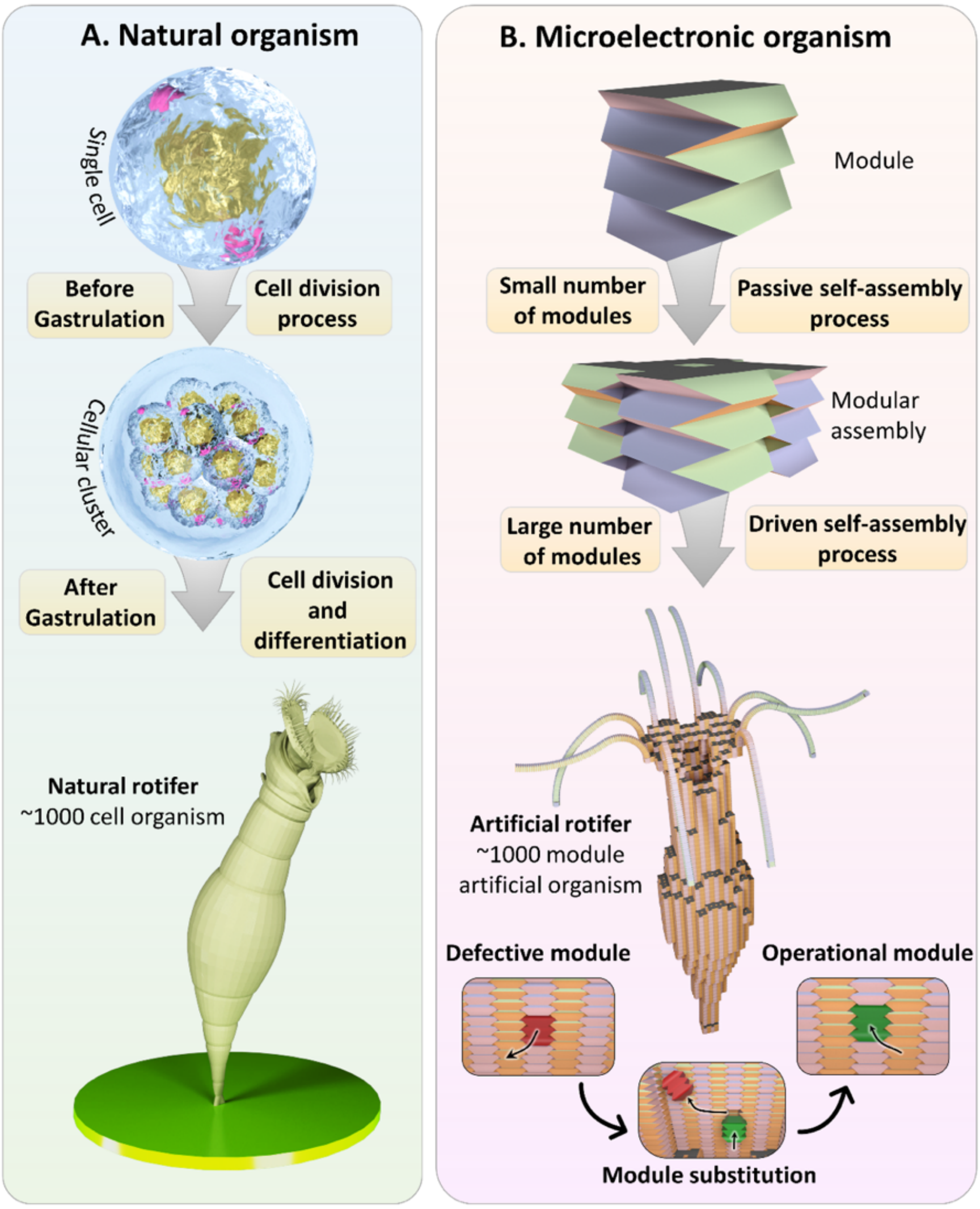

**Figure 7: Morphogenesis of natural and artificial organisms from their unit cells. A.** Example of a natural ~1000 cell multicellular organism, a rotifer, with morphogenesis via a gastrulating cell cluster **B.** Example of an artificial ~ 1000 smartlet organism, rotifer-like, with differentiated smartlet cells self-assembling under electronic control. The smartlet cell modules are folded up as structures with programmed flexibility that is retained during self-assembly via docking, while allowing electrical contacts to be made, giving the whole organism dynamic flexibility. The self-assembly at microscales would normally be slow, but the process can be radically accelerated compared with biological morphogenesis, by locomotion being directed (driven) by communication (direct or indirectly via a central processor) between smartlets with specific IDs. The self-assembly process, resulting in folded shells, also allows accessibility for the repair of the structure, replacing defective smartlets, as shown.

With this key functionality, SMARTLET differentiation variants simply depend on the choice of electronic signaling and regulation between SMARTLETs and the way in which electronic programs connect this information with electronic state changes, which may in turn result in changes to the portion of chiplet memory responsible for inheritable differences in operation (equivalent to gene expression control in cells). Of course, one should not argue for a one-to-one translation of cellular differentiation to SMARTLETs, as there are many differences and unique opportunities arising from the digital programmability of SMARTLETs – whether as serial or parallel machines. For microelectronic differentiation there are, as in any complex system that approaches biological complexity, fundamentally two ways to proceed, with specific or general solutions to the challenges posed: specific dynamical transitions in the electronic state control of electronic chiplets, depending on simple external trigger signals but complex multiply homeostatic dynamics, or more general transitions in electronic state controllers by changes to (somatically) inheritable information. Although this general mechanism is associated with some overhead in resources to establish, it can be used to allow module to module differentiation state transmission for any so-coupled dynamical system.

While this may all sound plausible with macroscopic electronic processors, this differentiated modular functionality will need to proceed down towards microscopic dimensions for microelectronic modular organisms to attain the illusion of apparently continuously varying complex materials. Simply put, scaling down module size leads to increased 3D resolution allowing more complex shapes to be assembled. Finer differentiation of modules allows more complex evolved devices to be constructed from a collection of modules working at finer scales. The main challenges in integrating self-assembled modules to organismic functions include: (i) distributed communication and energy management, (ii) intelligent coordination of distributed functions towards organismic goals, (iii) realization of global shape changes and locomotion (iv) global control of material transport including the transport of SMARTLET modules. The first three of these have already been discussed in previous sections, and while intensive work beyond the current considerations of macroscopic soft modular robotics will be required, considering the special capabilities of self-assembling electronic systems and the interplay of contacted and through-space communication, we foresee rapid realizations of these ideas in the coming years.

The global control of material transport in a self-assembling system, on the other hand, appears more far-fetched. Motorized SMARTLET modules can provide a 3D transport network, which together with shape changes in the form of opening and closing of tight holds can also enable the transport of fluid materials and concentrated substances in solution as cargo. Moreover, fine-grained SMARTLET modules can also potentially self-assemble microfluidic channels and chambers with active wall contractions and/or valves to permit the pipelined transport of fluids and suspended material. This can happen at two levels: either intra-SMARTLET with the alignment and sealing of through SMARTLET channels (see for example the "Swiss-roll" self-folding microfluidics in Figure 6) upon self-assembly, or via larger inter-SMARTLET channels formed by self-assemblies of SMARTLETs into shell-like structures forming channel walls (see Figure 6H). A prerequisite is the ability to form tight seals in the docking of self-assembled modules. A folded shell-layered architecture for microelectronic morphogenesis can even create (potentially fractal) 3D organisms in which each module is accessible for repair or exchange without having to pass through others.

Whether the number of different modules required to build organisms is limited or involves a large repertoire of combinatorial variants, provided their construction involves a fixed optimized set of established and automated clean room processes, there can be enormous economies of scale in production. Since SMARTLETs contain a small volume fraction of silicon chiplets, also amenable to volume foundry production, microelectronic morphogenesis is potentially an extremely economical approach to complex technological production and innovation via strong digitalization. Self-assembling SMARTLET organisms are also economical in terms of self-repair and recycling of components: with both active self-resorting and passive digital labelling assisted sorting machines enabled by the digital labelling of microscopic components.

The regulation of resources assigned to upholding electronic intercommunication pathways and the electronic transfer and assignment of energy to *smartlet* modules are all matters for self-organized negotiation, linked to overall operation in the artificial organisms. Of course, many such systems will prove exploitable and break down, but mechanisms can be put in place to reward useful function with energy and/or material nutrient supplies. Intelligence of the proposed artificial organisms will be discussed, including aspects like embodied intelligence, learning and the relationship between 2D genetic information underlying the morphology of the artificial organisms and their heredity and individual memory.

## 7. Outlook towards sustainable technology with microelectronic morphogenesis

The main scientific grand challenge that this review addresses is microelectronic artificial life, with the motivation "Life as we know it can only fully be understood in the context of life as it could be" [145]. While this paraphrase of Chris Langton's famous statement at the first workshop on Artificial Life[1], has been explored with some success in simulation, for example with digital organisms[146–149], the experimental study of life as it could be has so far only been convincingly alive very close to existing molecular life in fields like synthetic biology[150][151,152] and not yet even organic chemistry[153,154]. But are real living organisms made of very different substrates possible? Is our understanding misled by science fiction's robotic life, cyborgs, and fantastic life forms like Fred Hoyle's "Black Cloud"[155] and the nanotech swarms in Crichton's "Prey"[156]? The functionality of life poses formidable technological challenges that has limited progress with artificial life outside the computer in non-biological substrates. This seems to be about to change because of the newly pioneered technology reviewed in this article, that is potentially able to distribute nanometer scale structuring and electronic circuits into self-assembling 3D fold up architectures that can acquire significant autonomous functionality in space[76–78]. This review has addressed the question of whether this new technology is indeed powerful enough to create artificial organisms strongly decoupled from their biological substrate. The answer seems to be a qualified yes, provided we outsource the equivalent of the protein translation machinery, *i.e.,* the patterned fabrication of planar multilayer structures to separate human-controlled clean room fab lines where we both know how to manufacture the substrates for smart modules efficiently using the coded information embodied in the microelectronics of the deployed modules.

The artificial organisms discussed here directly address the standard suite of societal concerns about safety, responsibility and avoiding unwanted proliferation. Firstly, self-reproduction is dependent upon specific complex building blocks equipped with electronic circuits that can only be manufactured under human control in specialized high tech clean rooms – they are not available in the environment. Secondly, the CMOS chiplets on the building blocks not only allow unique machine-readable labels to be fabricated into them as an identifier providing complete information about their fabrication and source, but also allow their participation in

[1] which built on pioneering works by computer science founding figures John von Neumann[162] and Alan Turing[163],

artificial organisms to be recorded in the building blocks. Artificial organism assembly can thus be constrained to only use certain types of blocks and to record the organism identity in the block: this allows organisms self-assembling in complex environments to retain their identity and hence for the unambiguous assignment of production and operation responsibility. Different artificial organisms can in addition be built from a limited number of different generic types of building blocks allowing both economies of scale in mass production and optimization for a sustainable economy in reuse of components. The developmental programs for artificial organisms can be reprogrammed into the building blocks.

There is also an enormous need for microrobotic solutions to society's major practical challenges, especially in medicine but also in more sustainable and smarter manufacturing, environmental remediation, and other areas. This is evidenced by the recent explosion of publications in this area. Microelectronic morphogenesis can provide solutions where current machines fail: through rapid on-site assembly capabilities from microscopic components, through its soft and flexible active structure, through its fundamentally evolvable nature both at the level of building blocks and self-assembled structures, through its self-awareness and self-repair endowed by self-assembly-aware electronic circuitry, through its ability to communicate with and be controlled by human operators, through its ability to operate either as discrete motile swarms of microscopic agents or as complex multicellular organisms with emergent mechanical capabilities. These features are game changers in many application scenarios, and the intrinsically modular design will result in a general purpose, powerful and cost-effective technological solution platform. For example, the ability to access areas through microscopic openings in unassembled form and assemble remotely and quickly inside from components[157,158] is an enabler both for[159,160](minimally invasive surgery, dentistry, electroceuticals, tissue engineering) and for manufacturing and other machines (smart remote monitoring and microcontrol inside working machines, very rapid prototyping, next generation product labelling with active use recording, learning and feedback for product improvement). Another potential use is in environmental protection and agriculture as artificial organisms matching properties of natural organisms with identical structure to allow on site data collection on detailed environmental influences on organisms. As the technology develops, research collaborations already underway [161] with specialists from social, economic, legal, ethical, and environmental areas are key in ensuring beneficial, safe, and sustainable applications.

**Supporting Information**

Supporting Information is available from the author.

**Acknowledgements**

The presentation of the connections to Living Technology (a field cofounded by Author 1) have benefited from earlier discussions with Prof. Norman H. Packard.

# References

[1] M. A. Bedau, J. S. McCaskill, N. H. Packard, S. Rasmussen, *Artif Life* **2010**, *16*, 89.

[2] S. Singh, D. Chaudhary, A. D. Gupta, B. P. Lohani, P. K. Kushwaha, V. Bibhu, in *2022 3rd International Conference on Intelligent Engineering and Management (ICIEM)*, **2022**, pp. 447–454.

[3] Z. Yu, X. Shen, H. Yu, H. Tu, C. Chittasupho, Y. Zhao, *Pharmaceutics* **2023**, *15*, DOI 10.3390/pharmaceutics15030775.

[4] X. Mao, M. Liu, Q. Li, C. Fan, X. Zuo, *JACS Au* **2022**, *2*, 2381.

[5] H. İ. Dokuyucu, N. G. Özmen, *J Field Robot* **n.d.**, *n/a*, DOI https://doi.org/10.1002/rob.22139.

[6] F. Stella, J. Hughes, *Front Robot AI* **2023**, *9*, DOI 10.3389/frobt.2022.1059026.

[7] S. Zhang, X. Ke, Q. Jiang, Z. Chai, Z. Wu, H. Ding, *Advanced Materials* **2022**, *34*, 2200671.

[8] X. Dong, X. Luo, H. Zhao, C. Qiao, J. Li, J. Yi, L. Yang, F. J. Oropeza, T. S. Hu, Q. Xu, H. Zeng, *Soft Matter* **2022**, *18*, 7699.

[9] G.-Z. Yang, S. H. Collins, P. Dario, P. Fischer, K. Goldberg, C. Laschi, M. K. McNutt, *Sci Robot* **2021**, *6*, eabn2720.

[10] S. CARNOT, *Annales scientifiques de l'É.N.S.* **1872**, 393.

[11] I. Prigogine, *Science (1979)* **1978**, *201*, 777.

[12] A. Turing, *Philos Trans R Soc Lond B Biol Sci* **1952**, *237*, 37.

[13] A. T. Winfree, *J Chem Educ* **1984**, *61*, 661.

[14] S. Li, D. A. Matoz-Fernandez, A. Aggarwal, M. Olvera de la Cruz, *Proc Natl Acad Sci U S A* **2021**, *118*, DOI 10.1073/pnas.2025717118.

[15] L. S. PENROSE, *Ann Hum Genet* **1958**, *23*, 59.

[16] R. Jones, P. Haufe, E. Sells, P. Iravani, V. Olliver, C. Palmer, A. Bowyer, *Robotica* **2011**, *29*, 177.

[17] da V. Leonardo 1452-1519, *The Notebooks of Leonardo Da Vinci*, New York : Dover Publications, 1970., **1970**.

[18] I. Rechenberg, *Comput Methods Appl Mech Eng* **2000**, *186*, 125.

[19] W. Barthlott, M. Moosmann, I. Noll, M. Akdere, J. Wagner, N. Roling, L. Koepchen-Thomä, M. A. K. Azad, K. Klopp, T. Gries, *Philosophical Transactions of the Royal Society A* **2020**, *378*, 20190447.

[20] A. Ellery, *Biomimetics* **2020**, *5*, DOI 10.3390/biomimetics5030035.

[21] M. A. Bedau, J. S. McCaskill, N. H. Packard, S. Rasmussen, *Artif Life* **2010**, *16*, 89.

[22] A. Salehi-Reyhani, O. Ces, Y. Elani, *Exp Biol Med* **2017**, *242*, 1309.

[23] D. A. Hammer, N. P. Kamat, *FEBS Lett* **2012**, *586*, 2882.

[24] C. Hu, S. Pané, B. J. Nelson, *Annu Rev Control Robot Auton Syst* **2018**, *1*, 53.

[25] S. Yu, Y. Guo, H. Li, C. Lu, H. Zhou, L. Li, *ACS Appl Mater Interfaces* **2022**, *14*, 11989.

[26] A. Benayad, D. Diddens, A. Heuer, A. N. Krishnamoorthy, M. Maiti, F. le Cras, M. Legallais, F. Rahmanian, Y. Shin, H. Stein, M. Winter, C. Wölke, P. Yan, I. Cekic-Laskovic, *Adv Energy Mater* **2022**, *12*, 2102678.

[27] P. J. Wellmann, *Discov Mater* **2021**, *1*, 14.

[28] M. A. Trefzer, A. M. Tyrrell, *From Practice to Application. Springer* **2015**.

[29] S. Deepanjali, N. M. Sk, *Journal of Electronic Testing* **2022**, 1.

[30] G. S. Hornby, H. Lipson, J. B. Pollack, *Proc IEEE Int Conf Robot Autom* **2001**, *4*, 4146.

[31] S. Rasmussen, L. Chen, D. Deamer, D. C. Krakauer, N. H. Packard, P. F. Stadler, M. A. Bedau, *Science (1979)* **2004**, *303*, 963.

[32] T. Froese, N. Virgo, T. Ikegami, *Artif Life* **2014**, *20*, 55.

[33] L. Zhang, Z. Zhang, H. Weisbecker, H. Yin, Y. Liu, T. Han, Z. Guo, M. Berry, B. Yang, X. Guo, J. Adams, Z. Xie, W. Bai, *Sci Adv* **2023**, *8*, eade0838.

[34] K.-N. Chen, K.-N. Tu, *MRS Bull* **2015**, *40*, 219.

[35] Y. Dai, C. F. Xiang, Z. X. Liu, Z. L. Li, W. Y. Qu, Q. H. Zhang, *Applied Sciences (Switzerland)* **2022**, *12*, DOI 10.3390/app12020723.

[36] R. J. Alattas, S. Patel, T. M. Sobh, *Journal of Intelligent and Robotic Systems: Theory and Applications* **2019**, *95*, 815.

[37] D. Shah, B. Yang, S. Kriegman, M. Levin, J. Bongard, R. Kramer-Bottiglio, *Advanced Materials* **2021**, *33*, DOI 10.1002/adma.202002882.

[38] S. Doncieux, N. Bredeche, J.-B. Mouret, A. E. (Gusz) Eiben, *Front Robot AI* **2015**, *2*, 1.

[39] S. Doncieux, N. Bredèche, J.-B. Mouret, Eds., *New Horizons in Evolutionary Robotics*, Springer Berlin Heidelberg, Berlin, Heidelberg, **2011**.

[40] K. S. Pister, J. Kahn, B. E. Boser, *Smart Dust: BAA97-43 Proposal Abstract.*, Berkeley, USA, **1997**.

[41] V. K. Bandari, O. G. Schmidt, *Advanced Intelligent Systems* **2021**, *3*, 2000284.

[42] D. Seo, R. M. Neely, K. Shen, U. Singhal, E. Alon, J. M. Rabaey, J. M. Carmena, M. M. Maharbiz, *Neuron* **2016**, *91*, 529.

[43] A. Hirata, S. Kodera, K. Sasaki, J. Gomez-Tames, I. Laakso, A. Wood, S. Watanabe, K. R. Foster, *Phys Med Biol* **2021**, *66*, 08TR01.

[44] R. P. Feynman, in *APS Annual Meeting*, **1959**.

[45] N. Xia, G. Zhu, X. Wang, Y. Dong, L. Zhang, *Soft Matter* **2022**, DOI 10.1039/D2SM00891B.

[46] M. F. Reynolds, A. J. Cortese, Q. Liu, Z. Zheng, W. Wang, S. L. Norris, S. Lee, M. Z. Miskin, A. C. Molnar, I. Cohen, P. L. McEuen, *Sci Robot* **2022**, *7*, eabq2296.

[47] M. Urso, M. Pumera, *Adv Funct Mater* **2022**, *32*, 2112120.

[48] C. K. Schmidt, M. Medina-Sánchez, R. J. Edmondson, O. G. Schmidt, *Nat Commun* **2020**, *11*, 5618.

[49] J. Zhang, Z. Ren, W. Hu, R. H. Soon, I. C. Yasa, Z. Liu, M. Sitti, *Sci Robot* **2021**, *6*, DOI 10.1126/scirobotics.abf0112.

[50] Y. Li, M. Zhu, V. K. Bandari, D. D. Karnaushenko, D. Karnaushenko, F. Zhu, O. G. Schmidt, *Adv Energy Mater* **2022**, *12*, 2103641.

[51] V. V Zhirnov, R. K. Cavin III, *Microsystems for Bioelectronics: The Nanomorphic Cell*, William Andrew, **2010**.

[52] C. X. Du, H. A. Zhang, T. G. Pearson, J. Ng, P. L. McEuen, I. Cohen, M. P. Brenner, *Soft Matter* **2022**, *18*, 6404.

[53] J. Cui, T. Y. Huang, Z. Luo, P. Testa, H. Gu, X. Z. Chen, B. J. Nelson, L. J. Heyderman, *Nature* **2019**, *575*, 164.

[54] W. Palm, C. B. Thompson, *Nature* **2017**, *546*, 234.

[55] S. Brenner, R. A. Lerner, *Proceedings of the National Academy of Sciences* **1992**, *89*, 5381.

[56] B. Rivkin, C. Becker, F. Akbar, R. Ravishankar, D. D. Karnaushenko, R. Naumann, A. Mirhajivarzaneh, M. Medina-Sánchez, D. Karnaushenko, O. G. Schmidt, *Advanced Intelligent Systems* **2021**, *3*, 2000238.

[57] V. K. Bandari, Y. Nan, D. Karnaushenko, Y. Hong, B. Sun, F. Striggow, D. D. Karnaushenko, C. Becker, M. Faghih, M. Medina-Sánchez, *Nat Electron* **2020**, *3*, 172.

[58] C. Becker, B. Bao, D. D. Karnaushenko, V. K. Bandari, B. Rivkin, Z. Li, M. Faghih, D. Karnaushenko, O. G. Schmidt, *Nat Commun* **2022**, *13*, 2121.

[59] C. Becker, D. Karnaushenko, T. Kang, D. D. Karnaushenko, M. Faghih, A. Mirhajivarzaneh, O. G. Schmidt, *Sci Adv* **2019**, *5*, eaay7459.

[60] H. A. Bunzel, J. L. R. Anderson, A. J. Mulholland, *Curr Opin Struct Biol* **2021**, *67*, 212.

[61] V. A. Bespalov, N. A. Dyuzhev, V. Y. Kireev, *Nanobiotechnology Reports* **2022**, *17*, 24.

[62] M. Meloni, J. Cai, Q. Zhang, D. Sang-Hoon Lee, M. Li, R. Ma, T. E. Parashkevov, J. Feng, *Advanced Science* **2021**, *8*, 2000636.

[63] S. Pandey, M. Ewing, A. Kunas, N. Nguyen, D. H. Gracias, G. Menon, *Proceedings of the National Academy of Sciences* **2011**, *108*, 19885.

[64] K. S. Kwok, Q. Huang, M. Mastrangeli, D. H. Gracias, *Adv Mater Interfaces* **2019**, *1901677*, DOI 10.1002/admi.201901677.

[65] N. Lazarus, C. D. Meyer, S. S. Bedair, G. A. Slipher, I. M. Kierzewski, *ACS Appl Mater Interfaces* **2015**, *7*, 10080.

[66] D. Karnaushenko, T. Kang, O. G. Schmidt, *Adv Mater Technol* **2019**, *4*, 1800692.

[67] D. Karnaushenko, N. Münzenrieder, D. D. Karnaushenko, B. Koch, A. K. Meyer, S. Baunack, L. Petti, G. Tröster, D. Makarov, O. G. Schmidt, *Advanced Materials* **2015**, *27*, 6797.

[68] C. Becker, B. Bao, D. D. D. Karnaushenko, V. K. Bandari, B. Rivkin, Z. Li, M. Faghih, D. D. D. Karnaushenko, O. G. Schmidt, *Nat Commun* **2022**, *13*, 2121.

[69] C. N. Saggau, F. Gabler, D. D. Karnaushenko, D. Karnaushenko, L. Ma, O. G. Schmidt, *Advanced Materials* **2020**, *32*, 2003252.

[70] R. Huber, F. Kern, D. D. Karnaushenko, E. Eisner, P. Lepucki, A. Thampi, A. Mirhajivarzaneh, C. Becker, T. Kang, S. Baunack, B. Büchner, D. Karnaushenko, O. G. Schmidt, A. Lubk, *Nat Commun* **2022**, *13*, 3220.

[71] F. Catania, H. de Souza Oliveira, P. Lugoda, G. Cantarella, N. Münzenrieder, *J Phys D Appl Phys* **2022**, *55*, 323002.

[72] D. Karnaushenko, T. Kang, V. K. Bandari, F. Zhu, O. G. Schmidt, *Advanced Materials* **2020**, *32*, 1902994.

[73] D. Karnaushenko, T. Kang, O. G. Schmidt, *Adv Mater Technol* **2019**, *4*, 1800692.

[74] Y. Luo, M. R. Abidian, J.-H. Ahn, D. Akinwande, A. M. Andrews, M. Antonietti, Z. Bao, M. Berggren, C. A. Berkey, C. J. Bettinger, J. Chen, P. Chen, W. Cheng, X. Cheng, S.-J. Choi, A. Chortos, C. Dagdeviren, R. H. Dauskardt, C. Di, M. D. Dickey, X. Duan, A. Facchetti, Z. Fan, Y. Fang, J. Feng, X. Feng, H. Gao, W. Gao, X. Gong, C. F. Guo, X. Guo, M. C. Hartel, Z. He, J. S. Ho, Y. Hu, Q. Huang, Y. Huang, F. Huo, M. M. Hussain, A. Javey, U. Jeong, C. Jiang, X. Jiang, J. Kang, D. Karnaushenko, A. Khademhosseini, D.-H. Kim, I.-D. Kim, D. Kireev, L. Kong, C. Lee, N.-E. Lee, P. S. Lee, T.-W. Lee, F. Li, J. Li, C. Liang, C. T. Lim, Y. Lin, D. J. Lipomi, J. Liu, K. Liu, N. Liu, R. Liu, Y. Liu, Y. Liu, Z. Liu, Z. Liu, X. J. Loh, N. Lu, Z. Lv, S. Magdassi, G. G. Malliaras, N. Matsuhisa, A. Nathan, S. Niu, J. Pan, C. Pang, Q. Pei, H. Peng, D. Qi, H. Ren, J. A. Rogers, A. Rowe, O. G. Schmidt, T. Sekitani, D.-G. Seo, G. Shen, X. Sheng, Q. Shi, T. Someya, Y. Song, E. Stavrinidou, M. Su, X. Sun, K. Takei, X.-M. Tao, B. C. K. Tee, A. V.-Y. Thean, T. Q. Trung, C. Wan, H. Wang, J. Wang, M. Wang, S. Wang, T. Wang, Z. L. Wang, P. S. Weiss, H. Wen, S. Xu, T. Xu, H. Yan, X. Yan, H. Yang, L. Yang, S. Yang, L. Yin, C. Yu, G. Yu, J. Yu, S.-H. Yu, X. Yu, E. Zamburg, H. Zhang, X. Zhang, X. Zhang, X. Zhang, Y. Zhang, Y. Zhang, S. Zhao, X. Zhao, Y. Zheng, Y.-Q. Zheng, Z. Zheng, T. Zhou, B. Zhu, M. Zhu, R. Zhu, Y. Zhu, Y. Zhu, G. Zou, X. Chen, *ACS Nano* **2023**, *17*, 5211.

[75] Z. Qu, M. Zhu, Y. Yin, Y. Huang, H. Tang, J. Ge, Y. Li, D. D. Karnaushenko, D. Karnaushenko, O. G. Schmidt, *Adv Energy Mater* **2022**, *12*, 2200714.

[76] C. Becker, B. Bao, D. D. Karnaushenko, V. K. Bandari, B. Rivkin, Z. Li, M. Faghih, D. Karnaushenko, O. G. Schmidt, *Nat Commun* **2022**, *13*, 1.

[77] Q. Huang, B. Tang, J. C. Romero, Y. Yang, S. K. Elsayed, G. Pahapale, T.-J. Lee, I. E. Morales Pantoja, F. Han, C. Berlinicke, T. Xiang, M. Solazzo, T. Hartung, Z. Qin, B. S. Caffo, L. Smirnova, D. H. Gracias, *Sci Adv* **2022**, *8*, eabq5031.

[78] B. H. Kim, K. Li, J.-T. Kim, Y. Park, H. Jang, X. Wang, Z. Xie, S. M. Won, H.-J. Yoon, G. Lee, W. J. Jang, K. H. Lee, T. S. Chung, Y. H. Jung, S. Y. Heo, Y. Lee, J. Kim, T. Cai, Y. Kim, P. Prasopsukh, Y. Yu, X. Yu, R. Avila, H. Luan, H. Song, F. Zhu, Y. Zhao, L. Chen, S. H. Han, J. Kim, S. J. Oh, H. Lee, C. H. Lee, Y. Huang, L. P. Chamorro, Y. Zhang, J. A. Rogers, *Nature* **2021**, *597*, 503.

[79] T. Deng, C. Yoon, Q. Jin, M. Li, Z. Liu, D. H. Gracias, *Appl Phys Lett* **2015**, *106*, 203108.

[80] B. Bao, B. Rivkin, F. Akbar, D. D. Karnaushenko, V. K. Bandari, L. Teuerle, C. Becker, S. Baunack, D. Karnaushenko, O. G. Schmidt, *Advanced Materials* **2021**, *33*, 2101272.

[81] F. Akbar, B. Rivkin, A. Aziz, C. Becker, D. D. Karnaushenko, M. Medina-Sánchez, D. Karnaushenko, O. G. Schmidt, *Sci Adv* **2021**, *7*, eabj0767.

[82] F. Akbar, B. Rivkin, A. Aziz, C. Becker, D. D. Karnaushenko, M. Medina-Sánchez, D. Karnaushenko, O. G. Schmidt, *Sci Adv* **2021**, *7*, 1.

[83] B. Rivkin, C. Becker, B. Singh, A. Aziz, F. Akbar, A. Egunov, D. D. Karnaushenko, R. Naumann, R. Schäfer, M. Medina-Sánchez, D. Karnaushenko, O. G. Schmidt, *Sci Adv* **2021**, *7*, eabl5408.

[84] K. Malachowski, M. Jamal, Q. Jin, B. Polat, C. J. Morris, D. H. Gracias, *Nano Lett* **2014**, *14*, 4164.

[85] K. Krishnan, T. Tsuruoka, C. Mannequin, M. Aono, *Advanced Materials* **2016**, *28*, 640.

[86] D. Liu, C. Poon, K. Lu, C. He, W. Lin, *Nat Commun* **2014**, *5*, 4182.

[87] Y. Zhang, J. L. Sargent, B. W. Boudouris, W. A. Phillip, *J Appl Polym Sci* **2015**, *132*, 41683.

[88] M. Liu, L. Zhang, T. Wang, *Chem Rev* **2015**, *115*, 7304.

[89] Z. Fang, H. Song, Y. Zhang, B. Jin, J. Wu, Q. Zhao, T. Xie, *Matter* **2020**, *2*, 1187.

[90] L. H. Dudte, E. Vouga, T. Tachi, L. Mahadevan, *Nat Mater* **2016**, *15*, 583.

[91] C. L. Randall, E. Gultepe, D. H. Gracias, *Trends Biotechnol* **2012**, *30*, 138.

[92] S. Xu, Z. Yan, K. K.-I. K. Jang, W. Huang, H. Fu, J. Kim, Z. Wei, M. Flavin, J. Mccracken, R. Wang, A. Badea, Y. Liu, D. Xiao, G. Zhou, J. Lee, H. U. Chung, H. Cheng, W. Ren, A. Banks, X. Li, U. Paik, R. G. Nuzzo, Y. Huang, Y. Zhang, J. A. Rogers, *Science (1979)* **2015**, *347*, 154.

[93] A. I. Egunov, Z. Dou, D. D. Karnaushenko, F. Hebenstreit, N. Kretschmann, K. Akgün, T. Ziemssen, D. Karnaushenko, M. Medina-Sánchez, O. G. Schmidt, *Small* **2021**, *17*, 2002549.

[94] Q. Huang, T. Deng, W. Xu, C. Yoon, Z. Qin, Y. Lin, T. Li, Y. Yang, M. Shen, S. M. Thon, J. B. Khurgin, D. H. Gracias, *Advanced Intelligent Systems* **2023**, *5*, DOI 10.1002/aisy.202000195.

[95] J.-H. H. Cho, M. D. Keung, N. Verellen, L. Lagae, V. V. Moshchalkov, P. Van Dorpe, D. H. Gracias, *Small* **2011**, *7*, 1943.

[96] W. Lee, Y. Liu, Y. Lee, B. K. Sharma, S. M. Shinde, S. D. Kim, K. Nan, Z. Yan, M. Han, Y. Huang, Y. Zhang, J.-H. Ahn, J. A. Rogers, *Nat Commun* **2018**, *9*, 1417.

[97] S. Takamatsu, B. K. Nguyen, E. Iwase, K. Matsumoto, I. Shimoyama, in *2007 IEEE 20th International Conference on Micro Electro Mechanical Systems (MEMS)*, IEEE, **2007**, pp. 719–722.

[98] B. Rivkin, C. Becker, F. Akbar, R. Ravishankar, D. D. Karnaushenko, R. Naumann, A. Mirhajivarzaneh, M. Medina-Sánchez, D. Karnaushenko, O. G. Schmidt, *Advanced Intelligent Systems* **2021**, *3*, 2000238.

[99] B. Rivkin, C. Becker, B. Singh, A. Aziz, F. Akbar, A. Egunov, D. D. Karnaushenko, R. Naumann, R. Schäfer, M. Medina-Sánchez, D. Karnaushenko, O. G. Schmidt, *Sci Adv* **2021**, *7*, DOI 10.1126/sciadv.abl5408.

[100] M. Z. Miskin, A. J. Cortese, K. Dorsey, E. P. Esposito, M. F. Reynolds, Q. Liu, M. Cao, D. A. Muller, P. L. McEuen, I. Cohen, *Nature* **2020**, *584*, 557.

[101] V. K. Bandari, Y. Nan, D. Karnaushenko, Y. Hong, B. Sun, F. Striggow, D. D. Karnaushenko, C. Becker, M. Faghih, M. Medina-Sánchez, F. Zhu, O. G. Schmidt, *Nat Electron* **2020**, *3*, 172.

[102] X. Kuang, S. Wu, Q. Ze, L. Yue, Y. Jin, S. M. Montgomery, F. Yang, H. J. Qi, R. Zhao, *Advanced Materials* **2021**, *33*, 1.

[103] J. Zhang, R. H. Soon, Z. Wei, W. Hu, M. Sitti, *Advanced Science* **2022**, *9*, 2203730.

[104] Y. Li, M. Zhu, V. K. Bandari, D. D. Karnaushenko, D. Karnaushenko, F. Zhu, O. G. Schmidt, *Adv Energy Mater* **2022**, *12*, 2103641.

[105] Y. Li, M. Zhu, D. D. Karnaushenko, F. Li, J. Qu, J. Wang, P. Zhang, L. Liu, R. Ravishankar, V. K. Bandari, H. Tang, Z. Qu, F. Zhu, Q. Weng, O. G. Schmidt, *Nanoscale Horiz* **2022**, *8*, 127.

[106] J. M. Martel, K. C. Smith, M. Dlamini, K. Pletcher, J. Yang, M. Karabacak, D. A. Haber, R. Kapur, M. Toner, *Sci Rep* **2015**, *5*, 1.

[107] J. Huang, P. Yu, M. Liao, X. Dong, J. Xu, J. Ming, D. Bin, Y. Wang, F. Zhang, Y. Xia, *Sci Adv* **2023**, *9*, eadf3992.

[108] M. Li, M. Dixit, R. Essehli, C. J. Jafta, R. Amin, M. Balasubramanian, I. Belharouak, *Advanced Science* **2023**, 2300920.

[109] J.-H. Lee, G. Yang, C.-H. Kim, R. Mahajan, S.-Y. Lee, S.-J. Park, *Energy Environ Sci* **2022**.

[110] A. J. Cortese, C. L. Smart, T. Wang, M. F. Reynolds, S. L. Norris, Y. Ji, S. Lee, A. Mok, C. Wu, F. Xia, N. I. Ellis, A. C. Molnar, C. Xu, P. L. McEuen, *Proceedings of the National Academy of Sciences* **2020**, *117*, 9173.

[111] J. S. Randhawa, L. N. Kanu, G. Singh, D. H. Gracias, *Langmuir* **2010**, *26*, 12534.

[112] D. H. Gracias, J. Tien, T. L. Breen, C. Hsu, G. M. Whitesides, *Science (1979)* **2000**, *289*, 1170.

[113] M. Boncheva, D. H. Gracias, H. O. Jacobs, G. M. Whitesides, *Proceedings of the National Academy of Sciences* **2002**, *99*, 4937.

[114] M. Boyvat, M. Sitti, *Advanced Science* **2021**, *8*, 2101198.

[115] W. Kitana, I. Apsite, J. Hazur, A. R. Boccaccini, L. Ionov, *Adv Mater Technol* **2023**, *8*, 2200429.

[116] J. Nie, Q. Gao, J. Qiu, M. Sun, A. Liu, L. Shao, J. Fu, P. Zhao, Y. He, *Biofabrication* **2018**, *10*, 035001.

[117] L. J. Y. Ong, T. Ching, L. H. Chong, S. Arora, H. Li, M. Hashimoto, R. DasGupta, P. K. Yuen, Y.-C. Toh, *Lab Chip* **2019**, *19*, 2178.

[118] C. Liu, Q. Lin, H. Kim, M. Yim, *Auton Robots* **2023**, *47*, 211.

[119] J. Santoso, C. D. Onal, *Soft Robot* **2021**, *8*, 371.

[120] J. Zhang, J. Shi, J. Huang, Q. Wu, Y. Zhao, J. Yang, H. Rajabi, Z. Wu, H. Peng, J. Wu, *Advanced Intelligent Systems* **2023**, *2300048*, 2300048.

[121] R. Niu, C. X. Du, E. Esposito, J. Ng, M. P. Brenner, P. L. McEuen, I. Cohen, *Proc Natl Acad Sci U S A* **2019**, *116*, 24402.

[122] H. Li, F. Peng, X. Yan, C. Mao, X. Ma, D. A. Wilson, Q. He, Y. Tu, *Acta Pharm Sin B* **2022**, DOI https://doi.org/10.1016/j.apsb.2022.10.010.

[123] S. Y. Jung, J. S. Park, M. Kim, H. W. Jang, B. C. Lee, S. Baek, *Journal of Sensor Science and Technology* **2022**, *31*, 286.

[124] J. Joseph, B. Ma, B. T. Khuri-Yakub, *IEEE Trans Ultrason Ferroelectr Freq Control* **2022**, *69*, 456.

[125] D. Chen, X. Cui, Q. Zhang, D. Li, W. Cheng, C. Fei, Y. Yang, *Micromachines (Basel)* **2022**, *13*, 114.

[126] Z. Zhao, G. D. Spyropoulos, C. Cea, J. N. Gelinas, D. Khodagholy, *Sci Adv* **2022**, *8*, eabm7851.

[127] W. M. Jacobs, D. Frenkel, *J Am Chem Soc* **2016**, *138*, 2457.

[128] N. A. M. Araújo, R. A. da Costa, S. N. Dorogovtsev, J. F. F. Mendes, *Phys Rev Lett* **2018**, *120*, 188001.

[129] D. Rawtani, Y. K. Agrawal, *Nanobiomedicine* **2014**, *8*, 1.

[130] S. T. Ellison, S. Duraivel, V. Subramaniam, F. Hugosson, B. Yu, J. J. Lebowitz, H. Khoshbouei, T. P. Lele, M. Q. Martindale, T. E. Angelini, *Soft Matter* **2022**, *18*, 8554.

[131] S. Duraivel, V. Subramaniam, S. Chisolm, G. M. Scheutz, Brent. S. Sumerlin, T. Bhattacharjee, T. E. Angelini, *Biophys Rev* **2022**, *3*, 31307.

[132] Z. Gong, *Nanomaterials* **2021**, *11*, DOI 10.3390/nano11040842.

[133] A. Carlson, A. M. Bowen, Y. Huang, R. G. Nuzzo, J. A. Rogers, J. A. Rogers, A. Carlson, R. G. Nuzzo, A. M. Bowen, Y. Huang, *Adv. Mater.* **2012**, *24*, 5284.

[134] D. Stamenović, D. E. Ingber, *Soft Matter* **2009**, *5*, 1137.

[135] Q. Liu, W. Wang, M. F. Reynolds, M. C. Cao, M. Z. Miskin, T. A. Arias, D. A. Muller, P. L. McEuen, I. Cohen, *Sci Robot* **2021**, *6*.

[136] M. Z. Miskin, A. J. Cortese, K. Dorsey, E. P. Esposito, M. F. Reynolds, Q. Liu, M. Cao, D. A. Muller, P. L. McEuen, I. Cohen, *Nature 2020 584:7822* **2020**, *584*, 557.

[137] K. Patch, *Nat Biotechnol* **2021**, *39*, 255.

[138] D. A. Funke, P. Hillger, J. Oehm, P. Mayr, L. Straczek, N. Pohl, J. S. McCaskill, *IEEE Transactions on Circuits and Systems I: Regular Papers* **2017**, *64*, 3013.

[139] L. Xu, M. Lassiter, X. Wu, Y. Kim, J. Lee, M. Yasuda, M. Kawaminami, M. Miskin, D. Blaauw, D. Sylvester, in *2022 IEEE International Solid- State Circuits Conference (ISSCC)*, **2022**, pp. 1–3.

[140] C. Linghu, S. Zhang, C. Wang, J. Song, *npj Flexible Electronics* **2018**, *2*, 26.

[141] R. Fernandes, D. H. Gracias, *Adv Drug Deliv Rev* **2012**, *64*, 1579.

[142] W. Kitana, I. Apsite, J. Hazur, A. R. Boccaccini, L. Ionov, *Adv Mater Technol* **2023**, *8*, 2200429.

[143] J. S. McCaskill, G. von Kiedrowski, J. Oehm, P. Mayr, L. Cronin, I. Willner, A. Hermann, S. Rasmussen, F. Stepanek, N. H. Packard, **2012**.

[144] D. A. Funke, P. Mayr, L. Straczek, J. S. McCaskill, J. Oehm, N. Pohl, in *2016 IEEE International Conference on Electronics, Circuits and Systems (ICECS)*, IEEE, **2016**, pp. 512–515.

[145] C. G. Langton, in *Interdisciplinary Workshop on the Synthesis and Simulation of Living Systems* (Ed.: C. Langton), Addison-Wesley, Los Alamos, New Mexico, **1989**.

[146] W. Fontana, L. W. Buss, *Bull Math Biol* **1994**, *56*, 1.

[147] T. S. Ray, *Santa Fe* **1992**.

[148] G. Kruszewski, T. Mikolov, *Artif Life* **2022**, *27*, 277.

[149] C. G. Langton, *Physica D* **1986**, *22*, 120.

[150] K. Powell, *Nature* **2018**, *563*, 172.

[151] A. Y. Kharrazi, A. Zare, N. Chapin, S. Ghavami, A. Pandi, in *New Frontiers and Applications of Synthetic Biology* (Ed.: V. Singh), Academic Press, **2022**, pp. 83–101.

[152] A. Danchin, *Synth Biol* **2021**, *6*, ysab010.

[153] P. L. Luisi, *The Anatomical Record: An Official Publication of the American Association of Anatomists* **2002**, *268*, 208.

[154] M. Exterkate, A. J. M. Driessen, *ACS Omega* **2019**, *4*, 5293.

[155] F. Hoyle, *The Black Cloud*, Penguin UK, **2010**.

[156] M. Crichton, *Prey*, HarperCollins, **2002**.

[157] G. Gardi, S. Ceron, W. Wang, K. Petersen, M. Sitti, *Nat Commun* **2022**, *13*, 2239.

[158] J. Yu, B. Wang, X. Du, Q. Wang, L. Zhang, *Nat Commun* **2018**, *9*, 3260.

[159] J. Law, X. Wang, M. Luo, L. Xin, X. Du, W. Dou, T. Wang, G. Shan, Y. Wang, P. Song, X. Huang, J. Yu, Y. Sun, *Sci Adv* **2023**, *8*, eabm5752.

[160] J. Yu, D. Jin, K.-F. Chan, Q. Wang, K. Yuan, L. Zhang, *Nat Commun* **2019**, *10*, 5631.

[161] M. Arnold, D. Gesmann-Nuissl, T. Blaudeck, D. Karnaushenko, O. Schmidt, in *Smart Systems Integration (SSI) Conference*, Bruges, Belgium, **2023**, p. accepted.

[162] J. von Neumann, *Theory of Self-Reproducing Automata*, University Of Illinois Press, Illinois, USA, **1966**.

[163] A. M. Turing, *Philos Trans R Soc Lond B Biol Sci* **1952**, *237*, 37.

**Short Description**

Microelectronic morphogenesis, the creation and maintenance of complex functional structures by microelectronic information within shape-changing materials, is making large strides towards artificial electronic organisms. like genetic information. This article reviews the fundamental breakthroughs that enable inheritable electronic information to control morphology and acquire self-awareness as well as sustainable energy and material autonomy.

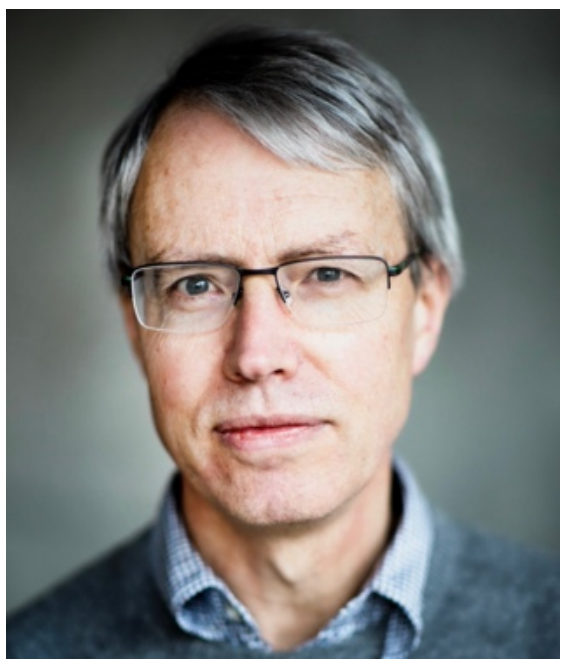

**John S. McCaskill** received his D. Phil. New College, Oxford University in 1983. He worked as group leader at the Max-Planck-Institute for Biophysical Chemistry, Göttingen with Manfred Eigen, then as full professor in 1992 at FSU Jena. In 1999 his department (BioMIP) moved to the Helmholtz Institute GMD, Sankt Augustin (merged with Fraunhofer, 2002), and 2004 to Ruhr University Bochum. Here he coordinated EU research into electronic microsystems for artificial life (PACE, ECCell, MICREAgents) and cofounded the European Centre for Living Technology in Venice. After work with a medical microrobotics startup Chemelion, he joined Chemnitz University of Technology, MAIN, 2022.

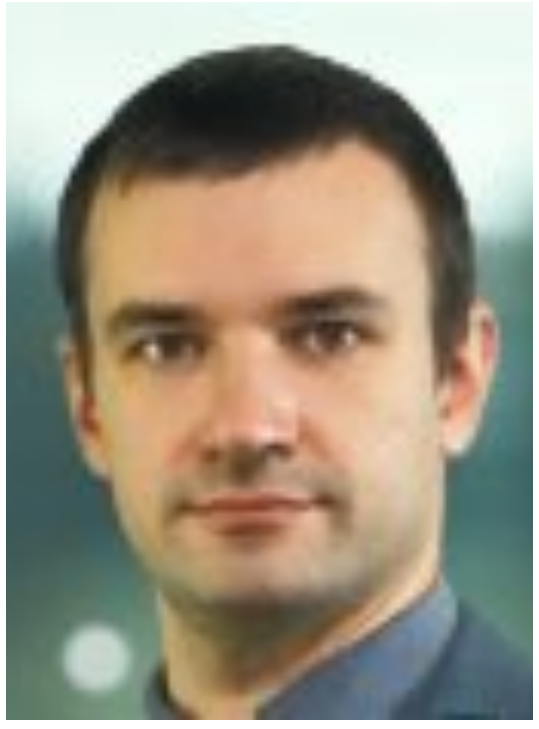

**Daniil Karnaushenko** received his Ph.D. from Chemnitz University of Technology in 2016 on Shapeable Microelectronics, following internships in Austria, Japan and USA. He has been a group leader at the Leibniz Institute for Solid State and Materials Research Dresden (Leibniz IFW Dresden). He subsequently moved to the Research Center for Materials, Architectures and Integration of Nanomembranes (MAIN) at the Chemnitz University of Technology, where he is a senior scientist and research group leader on shapeable and modular meso-electronics. His research focuses on novel microfabrication processes, including structural self-assembly techniques and thin-film microelectronics.

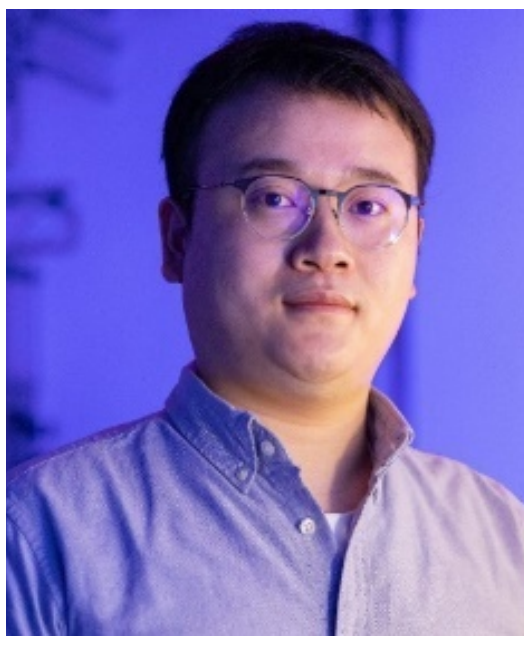

**Minshen Zhu** received the PhD degree from City University of Hong Kong, in 2017. He subsequently joined the Institute for Integrative Nanosciences at Leibniz IFW Dresden and led the research of energy storage at the microscale. In 2022, he moved to the Research Center for Materials, Architectures and Integration of Nanomembranes (MAIN) at Chemnitz University of Technology. Supported by the European Research Council (Starting Grant), his research activities aim to develop on-chip manufacturable dust-sized batteries for monolithic integration in intelligent microsystems.

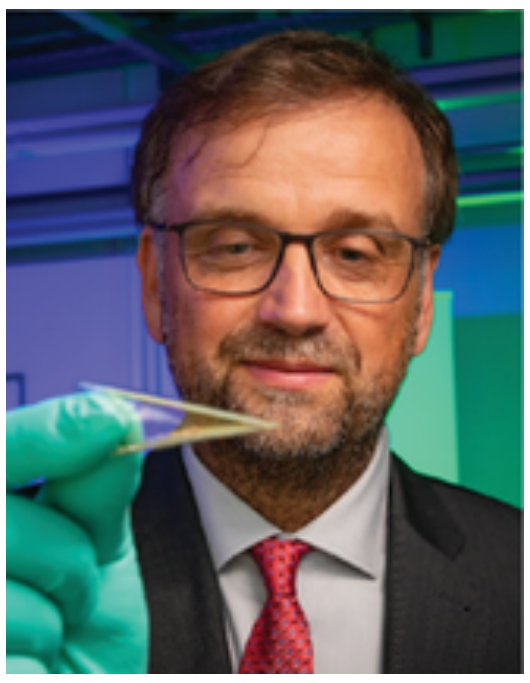

**Oliver G. Schmidt** received his Dr. rer. nat. degree from the TU Berlin in 1999. He holds a Full Professorship for Material Systems for Nanoelectronics and is the Scientific Director of the Research Center for Materials, Architectures and Integration of Nanomembranes (MAIN) at the Chemnitz University of Technology. He is an Adjunct Professor for Nanophysics at the TU Dresden and holds an Honorary Professorship at the Fudan University. His interdisciplinary activities bridge across several research fields, ranging from microrobotics and flexible electronics to microbatteries and biomedical applications.

# Supporting Information

**Microelectronic Morphogenesis: Progress towards Artificial Organisms**

*John S. McCaskill*, Daniil Karnaushenko*, Minshen Zhu and Oliver G. Schmidt**
*E-mail: john.mccaskill@main.tu-chemnitz.de ; daniil.karnaushenko@main.tu-chemnitz.de ; oliver.schmidt@main.tu-chemnitz.de

Three videos are available from the authors, but not on this platform because of document source constraints incurred by arXiv.

**Video S1 Folding self-assembly of SMARTLETs.**

SMARTLETs are self-assembled *via* a programmed material interaction, employing an environmental trigger that executes the process simultaneously and in parallel for all the wafer scale manufactured structures. The patterns and manufacturing parameters for individual layers, hinges, interconnects, sensors, and logic elements are encoded in digital "DNA", which is stored in an integrated chiplet. This information is reused in the manufacturing facility as a recipe for SMARTLETs fabrication allowing digital evolution of such smart mesoscale systems.

**Video S2 Animation of the hierarchical assembly process and organism movement.**

Individual SMARTLETs are sufficiently equipped to idle or operate autonomously for a limited amount of time during which the modules are intended to self-assemble via programmed interactions (*e.g.,* magnetic handshaking, surface tension etc.) forming sub-ohmic electrical connections. Assembled into an organism, modules can share their resources and operate together with other modules in a distributed controlled fashion, to locomote and conduct embodied energy management.

**Video S3 Modular self-repair by exchange of SMARTLETs in artificial organisms.**

The distributed controller can recognize the status of individual modules and trigger a substitution/repairment procedure to keep the overall modular organism operational.